\documentclass[10pt,journal,compsoc]{IEEEtran}
\ifCLASSOPTIONcompsoc
  \usepackage[nocompress]{cite}
\else
  \usepackage{cite}
\fi

\ifCLASSINFOpdf
\else
\fi

\usepackage{tcolorbox}
\usepackage{enumitem}
\tcbuselibrary{skins}

\usepackage{times}
\usepackage{epsfig}
\usepackage{graphicx}
\usepackage{amsmath}
\usepackage{amssymb}
\usepackage{footnote}
\usepackage{dsfont}
\usepackage{enumitem}

\usepackage{comment}
\usepackage{blindtext}
\usepackage{morewrites}
\usepackage{makecell}
\usepackage{pifont}
\usepackage{multirow}
\usepackage{xcolor}
\usepackage{float}
\usepackage{stfloats}

\usepackage[utf8]{inputenc} 
\usepackage[T1]{fontenc}    
\usepackage{hyperref}       
\usepackage{url}            
\usepackage{booktabs}       
\usepackage{amsfonts}       
\usepackage{nicefrac}       
\usepackage{microtype}      
\usepackage{xcolor}         

\usepackage{wrapfig}
\usepackage{graphicx}
\usepackage{amsmath}
\usepackage{mathtools}
\usepackage{multirow}
\usepackage{makecell}
\usepackage{tabularx}
\usepackage{algorithm2e}
\usepackage{fancyvrb,xcolor}
\usepackage{nicematrix}
\usepackage{bbm}

\usepackage{algorithm2e}
\usepackage{algorithmic}
\usepackage{fancyvrb}
\usepackage{xcolor}
\usepackage{nicematrix}

\usepackage{tikz}
\usepackage{comment}
\usepackage{amsmath,amssymb} %
\usepackage{color}
\usepackage{enumitem}
\usepackage{amsthm}

\usepackage{multirow}
\usepackage{makecell}
\usepackage{amsmath}
\usepackage{capt-of}
\usepackage{tabularx}
\usepackage{epsfig}
\usepackage{amssymb}
\usepackage{amsfonts}
\usepackage{booktabs}
\usepackage{scalerel}
\usepackage{listings}
\usepackage{varwidth}
\usepackage[export]{adjustbox}
\usepackage{tikz}
\usetikzlibrary{tikzmark}

\usepackage{stmaryrd}
\usepackage{bbm}
\usepackage{wrapfig}
\usepackage{pifont}
\usepackage[utf8]{inputenc}

\definecolor{deepblue}{rgb}{0,0,0.5}
\definecolor{officeblue}{RGB}{0,102,204}
\definecolor{deepred}{rgb}{0.6,0,0}
\definecolor{deepgreen}{rgb}{0,0.5,0}
\definecolor{mybrickred}{RGB}{182,50,28}

\definecolor{fillcolor}{RGB}{216,217,252}

\newcommand{\etal}{\textit{et al.}}
\newcommand{\ie}{\textit{i.e.}}
\newcommand{\eg}{\textit{e.g.}}

\usepackage{pifont}
\usepackage[dvipsnames]{xcolor}
\newcommand{\cmark}{\textcolor{OliveGreen}{\ding{51}}}
\newcommand{\xmark}{\textcolor{red}{\ding{55}}}
\newcommand{\omark}{\textcolor{blue}{\ding{72}}}

\begin{document}

\title{
    PRISM-RAG: Multimodal Hypergraph Retrieval-Augmented Generation for Tobacco Product and Legislative Policy Reasoning
}

\author{
Manuel~Serna-Aguilera~\IEEEmembership{Student~Member,~IEEE},
Raegan~Anderes,
Page Dobbs, \\ 
Khoa~Luu~\IEEEmembership{Senior Member,~IEEE}
\IEEEcompsocitemizethanks{
\IEEEcompsocthanksitem Manuel Serna-Aguilera and Khoa Luu are with the Electrical Engineering \& Computer Science Department, University of Arkansas, Fayetteville, Arkansas 72701. E-mail: mserna@uark.edu, khoaluu@uark.edu. 
\IEEEcompsocthanksitem Raegan Anderes is with the Department of Health, Human Performance and Recreation, University of Arkansas, Fayetteville, Arkansas 72701. E-mail: rmandere@uark.edu.
\IEEEcompsocthanksitem Page Dobbs is with the Fay W. Boozman College of Public Health, University of Arkansas for Medical Sciences, Little Rock, Arkansas, 72205. Email: PDDobbs@uams.edu.
}
}
\date{}



\IEEEtitleabstractindextext{%
\begin{abstract}
The disambiguation of semantically similar statutory text across jurisdictions is a retrieval problem that existing methods do not solve. 
This inter-context conflict can steer generative models toward confidently produced answers grounded in \textit{topically relevant but jurisdictionally incorrect}, sources. 
Tobacco and nicotine control regulations vary across the United States, from federal, state, and municipal statutes and administrative codes that may share similar language. Thus, robust reasoning requires identifying \textit{which} jurisdiction's law governs a given product, not merely retrieving semantically similar or relevant text for a given query. This is critical because emerging nicotine products (e.g., e-cigarettes and oral nicotine products) may circumvent existing regulations due to product definitions, and governments, agencies, and public health researchers and practitioners struggle to know what products are subject to which regulation. 
State-of-the-art (SOTA) document retrieval-augmented generation (RAG) methods break down in this task, rife with inter-context conflict, for two reasons. First, they provide no mechanism to ground a query in a product image, and second, embedding-similarity or entity-based retrieval cannot distinguish a jurisdiction's statute from another jurisdiction's similar text. 
To expose this failure case and provide a benchmark for evaluating solutions to it, we introduce the \textbf{Nico}tine \textbf{P}roduct and \textbf{R}egulation \textbf{I}mage-and-Text \textbf{S}urveillance \textbf{M}ultimodal (\textbf{NicoPRISM}) dataset, developed by our team of computer scientists and public health researchers. NicoPRISM comprises 161,563 images from web and social media sources paired with seven structured captions, a curated knowledge base of product, health, and \textit{static} legislative documents spanning \textbf{13} US exemplar jurisdictions, and 1,495 validated question-answer pairs organized into two benchmark tasks: policy compliance QA and product knowledge QA. 
To address the jurisdictional disambiguation problem directly, we propose \textbf{PRISM-RAG}, a multimodal hypergraph RAG framework that constructs a compact hypergraph over images, captions, and entities from heterogeneous document data without any large language model calls at index time. PRISM-RAG grounds every query in a product image, then routes retrieval through a jurisdiction-aware context assembly mechanism that guarantees statutory text from the queried jurisdiction reaches the language model by construction, independent of embedding-space topology. PRISM-RAG retrieves passages from the correct jurisdiction in 93.9\% of policy compliance queries, a 48.6 percentage point advantage over standard RAG ($p<0.001$), while requiring zero large language model calls at index time and a single call at query time. Across metrics measuring ground-truth keyword retrieval, semantic similarity, jurisdiction retrieval accuracy, compliance label accuracy, and context coverage, several of which expose limitations in SOTA document RAG methods, PRISM-RAG is competitive with and outperforms current SOTA document RAG frameworks while minimizing LLM calls at both index and query time. Data, embeddings, captions, documents, question-answer pairs, and source code are publicly released to support further public health research; project website at \url{https://manuelserna.github.io/sch-tpami-website/}. A subset of NicoPRISM is hosted on \href{https://kaggle.com/datasets/b5e316b13840a4e6d7f88742c219b0710869f368d439915b9dfefec88c5a5b30}{Kaggle}.
\end{abstract}

}

\maketitle
\section{Introduction} \label{sec:introduction}
Disambiguating semantically similar statutory text across jurisdictions is a retrieval problem that no existing method solves, yet it is central to tobacco and nicotine product surveillance. Tobacco and nicotine product innovation is rapidly outpacing monitoring efforts, and governments and public health researchers struggle to keep pace as products evolve to circumvent regulation. Identifying a nicotine or tobacco product from an image, understanding its flavors, marketing claims, and determining which regulations apply in a given jurisdiction are tasks public health researchers and policymakers must perform regularly, yet no existing tool or dataset supports all three at once, at scale. This problem presents two compounding challenges. 
First, a product image must serve as the entry point into a document knowledge base spanning product, health, and legislative documents, that is, cross-modal grounding. Visual product attributes, \eg, ``cool mint,'' is described in marketing language, while regulatory statutes refer to legal constructs, \eg, ``characterizing flavor,'' ``oral nicotine product,'' ``state preemption clause,'' whose semantics vastly differ. 
Second, state and municipal statutes frequently share similar language, thus, robust document reasoning requires identifying \textit{which} jurisdiction's law applies to a given product. Merely retrieving semantically similar text, as we will show in this work, is not sufficient. 
Bridging both challenges at scale, across diverse jurisdictions and product types, is the central challenge this work addresses.

Existing tobacco and nicotine image datasets have made valuable progress \cite{murthy-tobacco-dataset-2024, vassey2024scalable-tobacco, chappa2024phad}. They remain limited, however, because they mainly support product classification or detection and do not explore connecting product images to document knowledge. Thus, critical questions about \textit{what a product is} or \textit{what law could apply to it} cannot be properly explored. Retrieval-augmented generation (RAG) offers a natural framework for grounding language model responses in external knowledge, however, state-of-the-art (SOTA) document RAG methods, rife with inter-context conflict, fail at this task for two reasons. 
First, such methods, with strong document understanding \cite{jimenez2024hipporag, luo2025hypergraphrag}, cannot process images as retrieval entry points and, with API calls to large-language model (LLM) services at index and query time, make indexing expensive where legislation changes frequently; PRISM-RAG's caption-grounded concept hyperedges instead provide this entry point at zero index-time LLM cost. Second, and more critically, embedding-similarity retrieval alone cannot distinguish a jurisdiction's statute from another jurisdiction's near-identical text; no existing RAG system incorporates a mechanism to disambiguate semantically similar language across jurisdictions. This produces dangerous false positives, where a response appears grounded and confident while citing the wrong jurisdiction's law. We show that PRISM-RAG's jurisdiction-aware design, a guarantee holding regardless of embedding-space topology, addresses this fail case, retrieving passages from the correct jurisdiction in 93.9\% of policy compliance queries, a 48.6 percentage point advantage over standard retrieval ($p<0.001$) in Sec. \ref{sec:experiments}.

\vspace{2mm}
\noindent
\textbf{Contributions of this Work}. 
There are \textit{five main contributions} in this work. \textit{First}, we introduce \textbf{Nico}tine \textbf{P}roduct and \textbf{R}egulation \textbf{I}mage-and-Text \textbf{S}urveillance \textbf{M}ultimodal (\textbf{NicoPRISM}) dataset, a large-scale, multimodal, expert-validated dataset for tobacco and nicotine product surveillance, comprising images, structured attribute captions, a static heterogeneous document knowledge base, and two benchmark QA tasks spanning product knowledge and policy compliance. \textit{Second}, as a stepping stone towards more robust image-legislation document understanding, we propose \textbf{PRISM-RAG}, a multimodal hypergraph RAG framework that bridges visual product content and a static legislative document collection via image-conditioned seed retrieval, concept hyperedge traversal, and jurisdiction-aware context assembly, with no LLM calls at index time. \textit{Third}, we contribute expert-in-the-loop benchmark design to evaluate PRISM-RAG. Question-Answer (QA) pairs were co-developed and validated directly by public health domain experts, ensuring benchmark tasks reflect real regulatory monitoring workflows. \textit{Fourth}, we propose comprehensive evaluation against standard document retrieval and SOTA document (and image) RAG baselines, including new retrieval-level jurisdiction accuracy (JA), context coverage (CC), response groundedness (RG), and efficiency metrics (wall time per query). \textit{Finally}, all images, metadata, embeddings, captions, documents, QA pairs, and PRISM-RAG source code are released to support future research; project website is \href{https://manuelserna.github.io/sch-tpami-website/}{linked here}. A subset of NicoPRISM is hosted on \href{https://kaggle.com/datasets/b5e316b13840a4e6d7f88742c219b0710869f368d439915b9dfefec88c5a5b30}{Kaggle}.

\section{Related Work} \label{sec:related-work}

In this section, we discuss data collection efforts and related uses of LLMs with respect to PRISM-RAG. Sec. \ref{subsec:related-datasets} covers the tobacco and nicotine analysis literature. Sec. \ref{subsec:lmms} discusses foundational LLMs and large multimodal models (LMMs).  Sec. \ref{subsec:rag} discusses retrieval-based reasoning, focusing on document retrieval.

\subsection{Tobacco and Nicotine Analysis} \label{subsec:related-datasets}
Tobacco and nicotine products are among the leading preventable causes of death globally. Despite this, monitoring the rapidly evolving landscape of new products and marketing strategies remains a challenge for public health researchers and policymakers \cite{noauthor_who_nodate}. Machine learning has been increasingly applied to tackle this problem, but prior work has used small sample datasets and limited task scopes. Past studies have analyzed global tobacco survey data \cite{kim2021machine} to identify associations between policy and tobacco risk, classify smoking-cessation habits, and predict relapse patterns \cite{perski2023classification}. More recent work \cite{chappa2024advanced, kong2023understanding, murthy2023influence} has turned to visual and multimodal signals on social media. Vassey \etal \cite{vassey2024scalable-tobacco} compiled 6,999 labeled Instagram images for e-cigarette detection. Murthy \etal \cite{murthy-tobacco-dataset-2024} detected vaping devices in 826 annotated TikTok images. The PHAD \cite{chappa2024phad} assembled a dataset of social media videos. Meanwhile, Lakatos \etal \cite{lakatos2024multimodal} demonstrated that deep learning can identify covert tobacco advertisements through multimodal analysis.

\subsection{Large Multimodal Models} \label{subsec:lmms}
Large language models (LLMs) have been used to detect product-related discourse on social media \cite{twitter-tobacco-llm-2025}. Yet, their potential to support complex downstream reasoning tasks, such as cross-referencing product attributes against jurisdiction-specific policy or characterizing marketing strategies across brands at scale, has not been explored. Furthermore, all prior datasets and methods are confined to classification or detection, leaving a large gap for tasks that demand deeper product understanding and policy reasoning. LLMs trace their origins to the attention mechanism \cite{vaswani2017attention}. The family of Generative Pre-trained Transformers (GPT) models \cite{brown2020languagemodelsfewshotlearners, yenduri2023gpt, openai2024gpt4technicalreport} have considerable reasoning capabilities thanks to instruction tuning \cite{ouyang2022traininglanguagemodelsfollow} and human feedback \cite{stiennon2022learningsummarizehumanfeedback}. Open-source models such as LLaMA \cite{touvron2023llamaopenefficientfoundation, touvron2023llama2openfoundation} and Vicuna \cite{peng2023instruction} have also come to dominate LLM research efforts. The highly successful series of open-source LLaVA models \cite{liu2023visualinstructiontuning, liu2024improvedbaselinesvisualinstruction} were tuned to align LLMs and visual encoders, with further work optimizing memory footprint \cite{zhang2025llavaminiefficientimagevideo}. More modern commercial models, such as GPT-5 \cite{singh2025openaigpt5card}, Gemini-3 \cite{gemmateam2025gemma3technicalreport}, or the Claude suite \cite{anthropic2024claude3}, can be accessed via their respective APIs.

\subsection{Retrieval-Augmented Generation} \label{subsec:rag}
While LLMs and LMMs by themselves have strong reasoning capabilities, they are susceptible to hallucination \cite{huang2025survey-hallu, liu2024survey-hallu}. A popular solution is to combine the power of LLMs and knowledge bases organized as document chunks, graphs, or hypergraphs. Han \etal's GraphRAG survey \cite{han2025graphrag} provides further insight into this problem. A ``standard'' RAG solution stores text document chunks and retrieves them via similarity scoring. The graph RAG family organizes documents into a knowledge base as a simple graph. GRAG \cite{hu2025graggraphretrievalaugmentedgeneration} constructs the optimal subgraph in linear time. LightRAG \cite{guo2025lightrag} is an efficient method for finding an optimal subgraph. Further graph RAG works, such as PathRAG \cite{chen2025pathrag}, HippoRAG \cite{jimenez2024hipporag}, and HippoRAG~2 \cite{gutierrez2025ragmemorynonparametriccontinual}, can find further relations but struggle to return fine details. HyperGraphRAG \cite{luo2025hypergraphrag} expands to connect entities using $n$-ary relations to model more complex relationships. Critically, all of the above graph RAG methods are text-only: they have no mechanism to ground a product image query in the document graph and thus cannot route from a visual product to complex documents, \eg, legislation. To our knowledge, PRISM-RAG is the first hypergraph RAG framework to incorporate visual product images as the entry point for legislation document retrieval.

\section{The NicoPRISM Dataset} \label{sec:dataset}

We discuss the motivation, scope, and challenges of NicoPRISM in Sec. \ref{subsec:overview}. 
Sec. \ref{subsec:data-collection} discusses the image data aspect of NicoPRISM and how it improves on past works in terms of scale and scope. 
Sec. \ref{subsec:data-prep} details data post-processing. 
Sec. \ref{subsec:captions} discusses the attribute captions for the images in the web set.
Sec. \ref{subsec:documents} details the document knowledge base, with particular focus on the legislative component. 
Sec. \ref{subsec:qa} provides details on the QA benchmark. 

\subsection{Overview and Motivation} \label{subsec:overview}

The rapid proliferation of tobacco and nicotine products, in particular nicotine pouches, poses challenges for the public health space. New products pushed to market may not comply with laws or may use marketing language that tries to circumvent legislation. Now, scale this problem of identifying products and checking compliance across multiple jurisdictions in the United States. Addressing this at scale requires a system that can identify a product from its image, characterize it at the attribute level, and determine whether it is subject to a specific jurisdiction's regulations, simultaneously.

NicoPRISM serves as our challenging evaluation environment, exposing the limits of existing retrieval and reasoning methods. We compare against document RAG methods rather than VLM-based RAG methods because the primary challenge in NicoPRISM is legislative document understanding, parsing, indexing, and retrieving jurisdiction-specific statutory text, and connecting to image captions, not visual recognition, which we demonstrate with high accuracy (see Sec. \ref{subsec:product-retrieval}). Thus, strong document understanding is the necessary component for policy compliance reasoning.
Three compounding challenges surface in this image-document connection setting. 
\textbf{(i)} Cross-modal grounding, where a product image must serve as the entry point into a document graph built from legislative text. 
\textbf{(ii)} Jurisdictional disambiguation, where very similar statutory language across law documents makes semantic similarity an unreliable signal.
\textbf{(iii)} Heterogeneous knowledge fusion, where product images, attribute captions, product knowledge documents, health literature, and legislative texts must be jointly indexed and retrieved within a single static knowledge base. 
PRISM-RAG is evaluated against this environment, and understanding these challenge properties is necessary to understand why SOTA document RAG methods fail and what the method must do differently.

The NicoPRISM dataset comprises 161,563 images from three distinct visual distributions, seven structured attribute captions for all web-scraped images, a curated knowledge base of product, health, and legislative documents spanning 13 exemplar US jurisdictions, and 1,495 expert-validated image-question-answer (or simply ``QA'') pairs, developed in close collaboration with two specialized public health researchers. 

\noindent
\textbf{NicoPRISM Dataset License and Release.} NicoPRISM is released under a \texttt{\textbf{CC~BY-NC~4.0}} research-only license that prohibits commercial use and explicitly restricts access by tobacco, nicotine, and cannabis industry actors and their affiliates. Images are \textit{not redistributed directly}; we release per-image metadata, labels, captions, and image encoder embeddings, along with a download script.

\subsection{Image Collection} \label{subsec:data-collection}

NicoPRISM draws images from three complementary sources, each capturing a distinct visual distribution that a deployed surveillance system must handle.

\noindent\textbf{Web-scraped images.} The web split was collected via the Apify platform \cite{apify-citation}, spanning 277 brands and 154,735 images across ten product types. It provides the most brand-diverse and visually controlled imagery, with clean product photography, consistent backgrounds, clear packaging, and readable text. This split forms the primary knowledge base for PRISM-RAG and all baselines.

\noindent\textbf{Social media video frames.} The TikTok and YouTube splits were sampled from the PHAD dataset \cite{chappa2024phad}. The TikTok split contains 5,667 frames across 39 brands; the YouTube split contains 1,161 frames across 12 brands. These splits introduce the challenge of a significant visual distribution shift from the web set. Products appear in-context with users, in motion, partially occluded, and under variable lighting in video frames. A method that retrieves correctly from clean catalog imagery may fail substantially on social media frames of the same product

\noindent\textbf{Social Media Release.} These splits are sourced from the PHAD \cite{chappa2024phad}, whose data handling terms govern their use. We do not release original video files, account identifiers, or video-level metadata; the NicoPRISM release includes only per-frame labels and embeddings.

\begin{figure*}[!ht]
    \centering
    \includegraphics[width=0.75\linewidth]{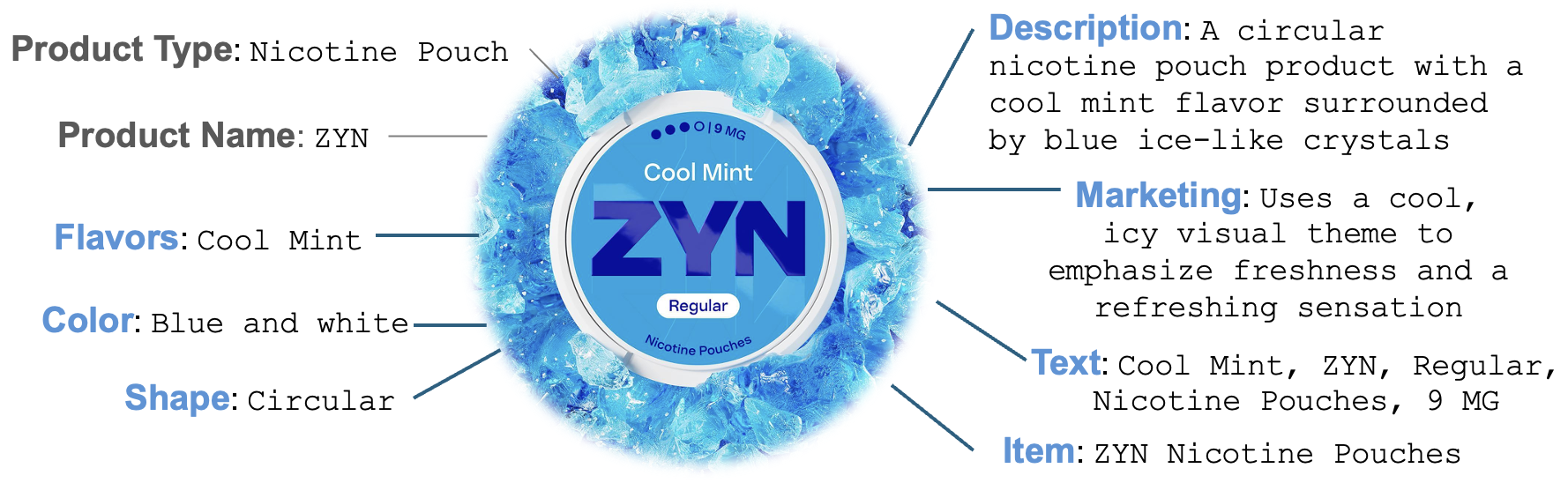}
    \caption{
        An image-caption sample from the NicoPRISM dataset of a nicotine pouch product with seven attribute captions.
        \textbf{Best viewed in zoom and in color.}
    }
    \label{fig:dataset-sample-and-doc-map}
\end{figure*}

\begin{figure*}[!t]
    \centering
    \includegraphics[width=0.7\linewidth]{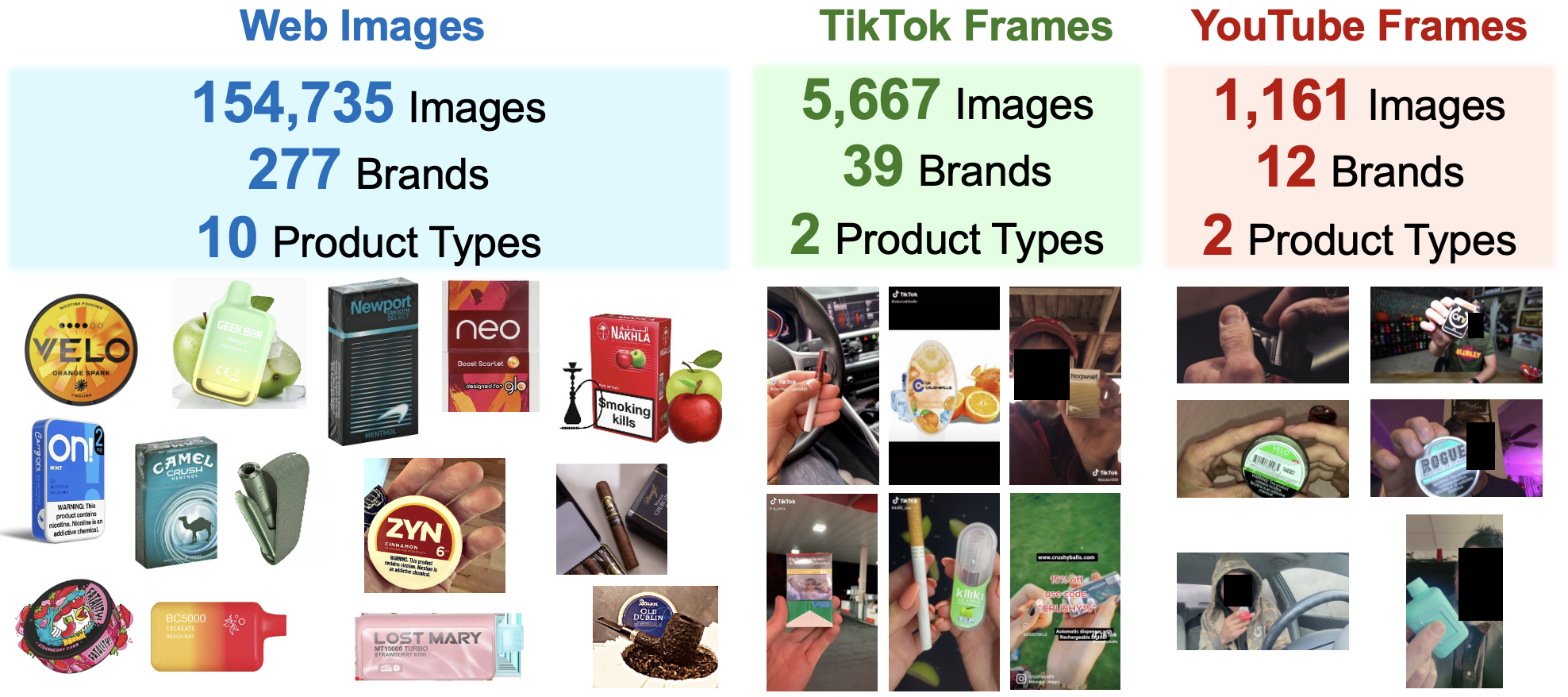}
    \caption{
        NicoPRISM Dataset statistics across the web set, TikTok set, and YouTube set, with image samples.
        \textbf{Best viewed in zoom and in color.}
    }
    \label{fig:dataset-stats1}
\end{figure*}

\noindent\textbf{Past Datasets} Table \ref{tab:tobacco-dataset-comparison} compares NicoPRISM with prior work along the dimensions relevant to these challenges; Figures \ref{fig:dataset-sample-and-doc-map} and \ref{fig:dataset-stats1} illustrate its primary components. Prior datasets are image-only and support only detection, providing no document-grounded knowledge base and no benchmark tasks that probe the failure modes above.
NicoPRISM is the first tobacco and nicotine product dataset to intentionally and jointly span all three challenge dimensions.

\begin{table*}[!t]\centering\small
\caption{
    A comparison of NicoPRISM with existing tobacco and nicotine product datasets.
    \textbf{Modality}: \texttt{I} = image only; \texttt{I+T+D} = image, text captions, and documents.
    \textbf{KB Types}: document categories in the knowledge base
    (Prod.\ = product knowledge, Health = health context, Legis.\ = legislative texts).
    Tasks: \texttt{DET} = detection, \texttt{RET} = image retrieval,
    \texttt{PKQ} = product knowledge QA, \texttt{PCQ} = policy compliance QA.
    \textbf{Expert Val.}\ = expert-validated QA pairs.
}\label{tab:tobacco-dataset-comparison}
\resizebox{\textwidth}{!}{
\begin{tabular}{lrrrcclrrc}\toprule
\textbf{Dataset}
    & \textbf{Images}
    & \textbf{Brands}
    & \textbf{Sources}
    & \textbf{Modality}
    & \textbf{Tasks}
    & \textbf{KB Types}
    & \textbf{Jurisdictions}
    & \textbf{QA Pairs}
    & \textbf{Expert Val.} \\\midrule
Murthy \etal \cite{murthy-tobacco-dataset-2024}
    & 826     & 3  & 1 & \texttt{I}
    & \{\texttt{DET}\} & -- & -- & -- & \xmark \\
Vassey \etal \cite{vassey2024scalable-tobacco}
    & 6,999   & 7  & 1 & \texttt{I}
    & \{\texttt{DET}\} & -- & -- & -- & \xmark \\
PHAD \cite{chappa2024phad}
    & 171,900 & 8  & 2 & \texttt{I}
    & \{\texttt{DET}\} & -- & -- & -- & \xmark \\
\midrule
\textbf{NicoPRISM} (Ours)
    & \textbf{161,563} & \textbf{277} & \textbf{3} & \texttt{I+T+D}
    & \{\texttt{RET}, \texttt{PKQ}, \texttt{PCQ}\}
    & Prod., Health, Legis.
    & \textbf{13} & \textbf{1,495} & \cmark \\
\bottomrule
\end{tabular}
}
\end{table*}

\subsection{Data Preparation and Expert Validation} \label{subsec:data-prep}

Mass scraping at the scale of NicoPRISM required a structured quality assurance pipeline. We employed a VLM to generate visual attribute captions for all images; keyword filtering of the resulting captions removed most irrelevant samples and enabled automated label reassignment where the depicted product differed from the scraped brand context. For TikTok and YouTube frames, team members individually inspected candidate frames and selected only those in which a tobacco or nicotine product was visible. This multi-stage pipeline (automated filtering followed by expert manual review) ensures that images associated with each policy compliance question accurately depict the product in question, a necessary precondition for valid compliance evaluation.

\subsection{Attribute Captions} \label{subsec:captions}

For every web-scraped image, we generate captions across seven structured attribute categories using a Qwen3.5-VL 4B VLM \cite{wu2025qwenimagetechnicalreport}: (i) item name, (ii) item description, (iii) flavors, (iv) marketing language, (v) shape, (vi) color, and (vii) visible text. 
These captions serve two purposes. First, they address the cross-modal grounding challenge directly; \ie, without a textual representation of the image's content, a text-based retrieval system has no surface to index or query. Secondly, the captions are structured around the attribute categories most relevant to legislative compliance. Notably, flavor characterization is the axis on which most flavor ban statutes turn, and enforcement bodies examine marketing language (and visible text) when determining whether a product is promoted in violation of youth marketing restrictions. Within the PRISM-RAG framework, captions define the $\mathcal{V}_A$ attribute nodes in the hypergraph and, through concept hyperedge construction, provide the principal semantic bridge from marketing language (\eg, ``cool mint'') to legislative terminology (\eg, ``characterizing flavor''). 

\subsection{Heterogeneous Knowledge Base Documents} \label{subsec:documents}

The document collection is the primary source of the jurisdictional disambiguation challenge and distinguishes NicoPRISM from prior product-image datasets. 
It is \textit{heterogeneous} in content type (product, health, legislative) and \textit{static} in scope. The document knowledge base is built once from publicly available sources and treated as a fixed document store at inference time, reflecting the realistic operational setting for a deployed surveillance tool where the system reasons from what it knows rather than looking up current law. We note that legislative statutes are not the only factor governing compliance status in practice; agency guidance documents, court rulings, and enforcement decisions can alter a product's effective legal standing independently of the underlying statute. Tracking the full legal landscape, including non-statutory policy instruments and judicial precedent, over time is beyond the current scope, and we treat the document collection as a static corpus reflecting the regulatory landscape as of March 2026.

\noindent\textbf{Product knowledge documents} include vendor catalog pages and encyclopedic articles, providing product-level factual context, ingredient composition, nicotine strength options, brand history, and flavor lineups.

\noindent\textbf{Health context documents} cover the known health effects and risks of tobacco and nicotine product use, drawing on clinical literature and government health agency articles.

\noindent\textbf{Legislative documents} are the component that poses the core retrieval challenge. We collected statutory texts from eight exemplar US states and five exemplar US cities, spanning the full spectrum of regulatory strictness.
The \textbf{\textit{eight states}} are: California, Massachusetts, and New Jersey (strict flavor or product bans); New York (relatively moderate); Illinois and Pennsylvania (moderate, with notable local variation); and Florida and Arkansas (lax or less restrictive, with preemptive state laws limiting local restrictions).
The \textbf{\textit{five cities}} are: San Francisco, CA; New York City, NY; Chicago, IL; Denver, CO; and Washington, D.C.
These jurisdictions were selected deliberately to represent the most consequential cases along the US regulatory spectrum and to capture legally complex preemption scenarios.
The challenge, therefore, is not merely scale but linguistic structure. That is, state and municipal ordinances frequently borrow language verbatim from federal statutes, making their embeddings nearly indistinguishable by cosine similarity. For instance, a system that retrieves by semantic similarity alone cannot reliably distinguish a D.C.-flavored tobacco ordinance from the federal Family Smoking Prevention and Tobacco Control Act when both use the phrase ``flavored tobacco product'' and define it identically. This is the specific failure mode that PRISM-RAG's jurisdiction-aware context assembly is designed to address.

\noindent\textbf{Scope.} All legislative documents reflect the regulatory landscape as of March 2026; statutory text changes over time, and users should verify the current status of any law referenced in system outputs.

\subsection{QA Benchmark} \label{subsec:qa}

The QA benchmark directly measures whether a method succeeds or fails across the three challenge dimensions defined above. We co-developed and validated it with our public health domain expert team.

NicoPRISM organizes the benchmark into two QA categories, each formulated as a set of image-conditioned questions paired with ground-truth keyword sets. We report the following evaluation metrics to capture as many aspects of the RAG process as possible in our challenging problem setting.
\textbf{KW-P}, keyword precision, quantifies how much of the final response $y^*$ is on-topic with the ground-truth keywords.
We also report \textbf{KW-R} (keyword recall) and \textbf{KW-F1} (F1 similarity score). 
\textbf{CC} (context coverage) is the fraction of ground-truth keywords present anywhere in the assembled \textit{context} $\mathcal{C}_q$, measuring \textit{retrieval quality} independently of generation quality. CC accounts for cases where a method retrieves the right content but generates poorly, or generates fluently from the wrong content. 
\textbf{RG} (response groundedness) is the cosine similarity between $y^*$ and $\mathcal{C}_q$, measuring how closely the response tracks the retrieved evidence. 
Following HyperGraphRAG \cite{luo2025hypergraphrag}, \textbf{Judge} is the LLM-as-a-Judge score (1--5) \cite{que2024hellobench}.
For policy compliance questions, we report \textbf{CA} (compliance accuracy), \ie, the fraction of responses whose LLM-judge-extracted compliance label matches the ground-truth labels, and \textbf{JA} (retrieval-level jurisdiction accuracy), \ie, the fraction of retrieved passages whose source document belongs to the queried jurisdiction.
JA directly measures the second challenge dimension: whether the retrieval mechanism is routing to the correct jurisdiction's documents at all, independent of the LLM stage. 
Critically, a method that scores low JA and moderate CA succeeds at compliance determination despite retrieving wrong-jurisdiction content; this gives insight into confident answers grounded in the wrong documents. 

\noindent\textbf{(I) Product knowledge QA} questions probe \textit{product-level attributes} including flavor profiles, marketing claims, packaging design, and nicotine strength framing. Questions range from direct attribute queries (\eg, ``What nicotine strengths does this product offer?'') to comparative queries requiring cross-product knowledge retrieval (\eg, ``Which other brands offer a flavor with this name?'').

\noindent\textbf{(II) Policy compliance QA} questions probe how US regulations apply to specific product types depicted in the query image. These questions span cases of graded legal complexity: from straightforward single-jurisdiction determinations (\eg, ``Is the sale of this flavored nicotine pouch legal in Arkansas as of 2024?'') to multi-hop scenarios requiring joint reasoning over state and municipal law (\eg, questions involving state preemption of a local flavor ban). Policy QA exercises all three challenge dimensions: cross-modal grounding (the query begins with an image and a question), jurisdictional disambiguation (the method must retrieve from the correct jurisdiction's statute, not a semantically similar one from another jurisdiction), and heterogeneous knowledge fusion (both product attribute evidence, captions, and legislative evidence may be relevant to the answer). Our expert collaborators constructed and finalized the ground-truth keyword sets by carefully revising legislative documents, ensuring that correct answers require actual statutory grounding. 

In total, NicoPRISM contains 1,495 distinct image-question pairs, comprising 170 product knowledge QA pairs and 1,325 policy compliance QA pairs. This expert-in-the-loop validation process is a key quality guarantee that distinguishes the NicoPRISM compliance benchmark from prior work.

\noindent
\textbf{Legal Scope Disclaimer.} PRISM-RAG outputs are intended to assist researchers and practitioners and do not constitute legal advice; users should consult qualified legal experts before acting on any system-generated policy compliance determination.

\section{The PRISM-RAG Framework} \label{sec:method}

\begin{figure*}[!t]
    \centering
    \includegraphics[width=0.8\linewidth]{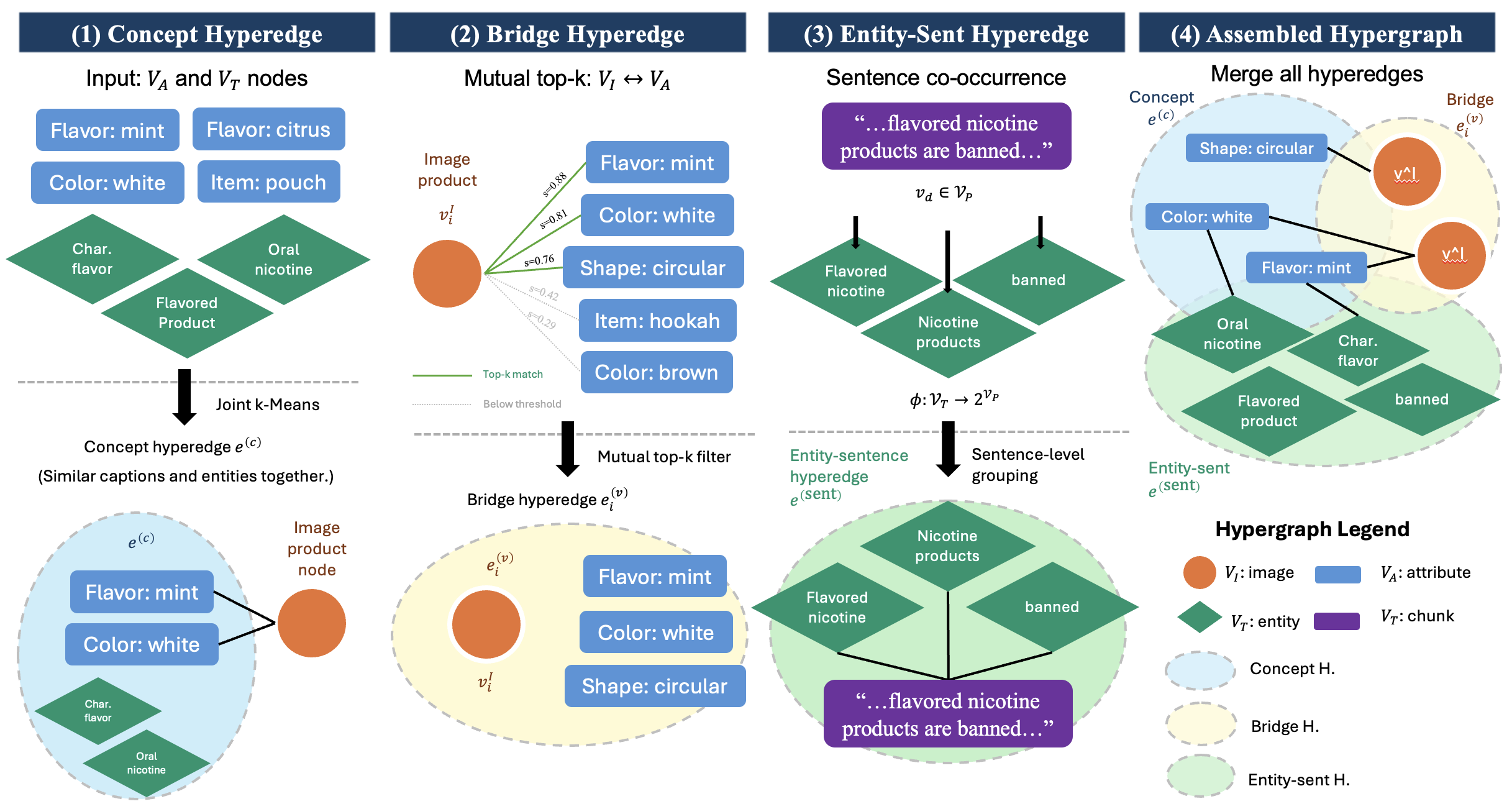}
    \caption{
        Our proposed PRISM-RAG framework is organized into a large-scale hypergraph for multimodal RAG.
        \textbf{(1)} \textbf{\textit{Concept hyperedges}} connect attributes and document entities.
        \textbf{(2)} Images to similar attributes form \textbf{\textit{bridge hyperedges}}.
        \textbf{(3)} \textbf{\textit{Entity-sentence hyperedges}} connect entities found within sentences within documents.
        \textbf{(4)} The assembled multimodal hypergraph combines nodes from steps (1--3) \textit{with no large model calls}.
        \textbf{Best viewed in color and with zoom.}
    }
    \label{fig:method}
\end{figure*}

In this section, we introduce PRISM-RAG, a multimodal RAG framework that connects image descriptions (\ie, captions) to heterogeneous documents; in this work, we focus on establishing a strong foundation for understanding and retrieving relevant legislative text. We discuss notation and data modeling assumptions for PRISM-RAG in Sec. \ref{subsec:data-modeling}. Sec. \ref{subsec:construction} details our indexing procedure. Sec. \ref{subsec:mm-query-procedure} describes the query procedure. Sec. \ref{subsec:response-gen} describes PRISM-RAG's response generation. Lastly, Sec. \ref{subsec:cost-theory} covers costs associated with PRISM-RAG, although index and retrieval complexity are similar across all SOTA RAG baselines.

\subsection{Data Modeling} \label{subsec:data-modeling}

We can denote the image dataset as $\mathcal{I} = \{ (x_i^{\mathcal{I}}, \mathcal{Y}_i, \mathcal{C}_i) \}^{N}_{i=1}$, where $\mathcal{Y}_i = (y_i^{(pn)}, y_i^{(pt)})$ and $\mathcal{C}_i = \{x_{i,l}^{\mathcal{C}}\}_{l=1}^{L}$. Image embeddings are $z_i^{\mathcal{I}} = f^{(\mathcal{I})}(x_i^{\mathcal{I}})$. Since caption-to-image alignment is required for bridge hyperedge construction (Sec. \ref{subsec:bridge}), we maintain two caption embeddings. \textit{Aligned} caption embeddings $\mathcal{Z}_i^{\mathcal{C}}$ are produced by the text branch $f^{(\mathcal{C})}$ of the same vision-language model used for $f^{(\mathcal{I})}$ (\eg, SigLIP \cite{zhai2023sigmoidlosslanguageimage}). \textit{Semantic} caption embeddings $\mathcal{Z}_i^{\mathcal{C},\mathcal{D}}$ are produced by a dedicated document encoder $f^{(\mathcal{D})}$ and are used for concept hyperedge clustering (Sec. \ref{subsec:concept}). The document set is $\mathcal{D}$, and each policy document $d \in \mathcal{D}$ carries a jurisdiction label $\mathrm{jur}(d) \in \mathcal{J}_{\mathrm{known}}$. 

\subsection{Construction} \label{subsec:construction}

\begin{table*}[!t]\centering\small
\caption{
    Role of each node set and hyperedge type across all phases of PRISM-RAG.
    \cmark = participates directly; \omark = participates passively; \xmark = does not participate.
}\label{tab:node-hyperedge-roles}
\resizebox{\textwidth}{!}{
\begin{tabular}{lcccccccc}\toprule
& \multicolumn{4}{c}{\textbf{Index Time}} & \multicolumn{4}{c}{\textbf{Query Time}} \\
\cmidrule(lr){2-5}\cmidrule(lr){6-9}
\textbf{Component} & \textbf{Bimodal Emb.} & \textbf{Concept HE} & \textbf{Bridge HE} & \textbf{Entity Extract.} & \textbf{Seed Retrieval} & \textbf{HE Scoring} & \textbf{Node Scoring $s(\mathbf{q},v)$} & \textbf{Context Assembly} \\
\midrule
\multicolumn{9}{l}{\textit{Node sets}} \\
$\mathcal{V}_I$ (image product)    & \cmark & \cmark & \cmark & \xmark & \cmark & \cmark & \cmark & \xmark \\
$\mathcal{V}_A$ (caption attribute) & \xmark & \cmark & \cmark & \xmark & \xmark & \cmark & \cmark & \xmark \\
$\mathcal{V}_T$ (document entity)  & \xmark & \cmark & \xmark & \cmark & \xmark & \cmark & \cmark & \omark \\
$\mathcal{V}_J$ (jurisdiction anchor) & \xmark & \xmark & \xmark & \omark & \xmark & \xmark & \xmark & \omark \\
$\mathcal{V}_P$ (document chunk)   & \xmark & \xmark & \xmark & \omark & \xmark & \omark & \cmark & \cmark \\
\midrule
\multicolumn{9}{l}{\textit{Hyperedge types}} \\
Concept $e^{(c)}$ & \xmark & \cmark & \xmark & \xmark & \xmark & \cmark & \xmark & \xmark \\
Bridge $e_i^{(v)}$ & \xmark & \xmark & \cmark & \xmark & \xmark & \cmark & \xmark & \xmark \\
Entity sentence $e^{(\text{sent})}$ & \xmark & \xmark & \xmark & \cmark & \xmark & \cmark & \xmark & \xmark \\
\bottomrule
\end{tabular}
}
\end{table*}

\subsubsection{Knowledge Definition}

We organize heterogeneous knowledge as a hypergraph $\mathcal{G} = (\mathcal{V}, \mathcal{E})$ with $\mathcal{V} = \mathcal{V}_I \cup \mathcal{V}_A \cup \mathcal{V}_P \cup \mathcal{V}_J \cup \mathcal{V}_T$. \textit{Image product nodes} $\mathcal{V}_I$ carry bimodal embeddings $\mathbf{h}_i^P$ (Eqn. \ref{eq:bimodal}). \textit{Caption attribute nodes} $\mathcal{V}_A$ encode unique (category, value) pairs with $f^{(\mathcal{D})}$. \textit{Document chunk nodes} $\mathcal{V}_P$ encode text chunks of all $d \in \mathcal{D}$ and carry jurisdiction metadata. \textit{Jurisdiction anchor nodes} $\mathcal{V}_J$ map jurisdiction name substrings to corresponding policy documents, enabling cost-effective jurisdiction hint detection without modifying chunk text. \textit{Document entity nodes} $\mathcal{V}_T$ are canonical entity strings from $d \in \mathcal{D}$, with source chunks recorded by $\phi: \mathcal{V}_T \to 2^{\mathcal{V}_P}$. Table \ref{tab:node-hyperedge-roles} summarizes each component's role.

\subsubsection{Multimodal Node Embeddings}

Each product node $v_i \in \mathcal{V}_I$ carries a bimodal embedding, as in Eqn. \eqref{eq:bimodal}, \ie, the $\ell_2$-normalized concatenation of the image and mean-pooled aligned caption representations.

\begin{equation}\label{eq:bimodal}
    \mathbf{h}_i^{P}
    = \left[\,
        \frac{z_i^{\mathcal{I}}}{\|z_i^{\mathcal{I}}\|_2}
        \;;\;
        \frac{\tilde{z}_i^{\mathcal{C}}}{\|\tilde{z}_i^{\mathcal{C}}\|_2}
      \,\right] \in \mathbb{R}^{n_{\mathcal{I}} + n_{\mathcal{C}}}
\end{equation}

Attribute nodes $v_k \in \mathcal{V}_A$, jurisdiction anchor nodes $v_j \in \mathcal{V}_J$, document entity nodes $v_t \in \mathcal{V}_T$, and document chunk nodes $v_d \in \mathcal{V}_P$ are all encoded with $f^{(\mathcal{D})}$. For bridge hyperedge construction, each attribute node also carries an aligned embedding $\bar{\mathbf{h}}_k^A \in \mathbb{R}^{n_\mathcal{C}}$.

\subsubsection{Document Entity Extraction and Document Node Construction}
\label{subsec:entity-extraction}

We extract named entities from all documents $d \in \mathcal{D}$ to form $\mathcal{V}_T$ via three different strategies.

\noindent\textbf{Cost-effective and LLM-free entity extraction (proposed).} We propose to apply a spaCy transformer-based NLP pipeline \footnote{\url{https://github.com/explosion/spaCy}}  to each document. Named entities and noun chunks are associated with source chunks via $\phi$. All entity nodes co-occurring within the same sentence are jointly connected as a single $n$-ary entity sentence hyperedge. This pipeline requires no LLM API calls to process document entities at index time.

\noindent\textbf{LLM-based extraction (ablation).} Following HyperGraphRAG \cite{luo2025hypergraphrag}, we prompt an LLM for entities and triples from each chunk, at the cost of one API call per document chunk.

\noindent\textbf{Ontology augmentation (ablation).} We optionally augment each legislative $\mathcal{V}_T$ entity's text with domain synonyms drawn from a hand-curated ontology before encoding, pulling entity embeddings toward the product attribute clusters they semantically belong to. 

\subsubsection{Concept Hyperedge Construction}
\label{subsec:concept}

\textit{Concept hyperedges}, $e^{(c)}$, defined in Eqn. \eqref{eq:concept-edge}, are formed by jointly clustering all $\mathcal{V}_A$ and $\mathcal{V}_T$ nodes in the $f^{(\mathcal{D})}$ embedding space, producing structures that bridge product attribute semantics and legislative entity semantics.

\begin{equation}
\label{eq:concept-edge}
\begin{split}
    e^{(c)} &= \mathcal{V}_A^{(c)} \;\cup\; \mathcal{V}_T^{(c)} \;\cup\; \\ &\Bigl\{\, v_i^I \;\Big|\; \exists\, v_k \in \mathcal{V}_A^{(c)} \cup \mathcal{V}_T^{(c)} : v_k \in \mathcal{N}(v_i^I) \,\Bigr\}
\end{split}
\end{equation}

For example, a cluster representing a cooling or menthol concept groups product nodes whose captions contain mint-adjacent attributes, with legislative entity nodes referring to ``menthol characterizing flavor,'' directly connecting products to the policies that govern such a product. We consider two clustering implementations of $\eta$: \textbf{(i) $k$-Means} with $C = 50$ clusters; and \textbf{(ii) Agglomerative Clustering} with cosine distance threshold $\delta_{\text{cluster}} = 0.30$. We test different values of $C \in \{50, 100, 150, 200\}$.

\subsubsection{Bridge Hyperedge Construction} \label{subsec:bridge}

\textit{Bridge hyperedges}, $e^{(v)}$ defined in Eqn. \eqref{eq:bridge-mutual}, connect each product image node $v_i^I \in \mathcal{V}_I$ to $\mathcal{V}_A$ attribute nodes whose aligned text embeddings are visually close to the image embedding. The bidirectionality constraint, implemented by Eqns. \eqref{eq:top-attrs-matches-for-img} and \eqref{eq:top-img-matches-for-attr}, reduces noise. We set $k = 10$ in all experiments. We also evaluate threshold-based bridge construction as an ablation ($\tau = 0.25$). With all three hyperedge types, $\mathcal{E} = \{e^{(c)}\}_{c=1}^{C} \cup \{e_i^{(v)}\}_{i=1}^{N} \cup \{e^{(\text{sent})}_d\}$.

\begin{align}
    \mathcal{T}_k^{\text{img}}(v_i^I) &= \operatorname*{top\text{-}k}_{v_k^A \in \mathcal{V}_A}\; {\mathbf{h}_i^{P,\mathcal{I}}}^{\!\top}\bar{\mathbf{h}}_k^{A} \label{eq:top-attrs-matches-for-img} \\
    \mathcal{T}_k^{\text{attr}}(v_k^A) &= \operatorname*{top\text{-}k}_{v_i^I \in \mathcal{V}_I}\; {\mathbf{h}_i^{P,\mathcal{I}}}^{\!\top}\bar{\mathbf{h}}_k^{A} \label{eq:top-img-matches-for-attr} \\
    \begin{split}
        e_i^{(v)} &= \{v_i^I\} \;\cup\; \\ &\Bigl\{\, v_k^A \;\Big|\; v_k^A \in \mathcal{T}_k^{\text{img}}(v_i^I), v_i^I \in \mathcal{T}_k^{\text{attr}}(v_k^A) \,\Bigr\} \label{eq:bridge-mutual}
    \end{split}
\end{align}

\subsection{Multimodal Query Procedure} \label{subsec:mm-query-procedure}

At query time, PRISM-RAG proceeds through four distinct steps. 

\noindent\textbf{Query Encoding.} Queries are encoded into a bimodal representation, given by Eqn. \eqref{eq:query-def}. Additionally, $\mathbf{q}^{\mathcal{D}} = f^{(\mathcal{D})}(q^{\mathcal{C}})/\|f^{(\mathcal{D})}(q^{\mathcal{C}})\|_2$ is used for scoring $\mathcal{V}_A$, $\mathcal{V}_T$, and $\mathcal{V}_P$ nodes. A jurisdiction hint $j_q$ is extracted by substring-matching known jurisdiction names against $q^{\mathcal{C}}$.

\begin{equation}\label{eq:query-def}
    \mathbf{q}
    = \left[\,
        \frac{f^{(\mathcal{I})}(q^{\mathcal{I}})}{\|f^{(\mathcal{I})}(q^{\mathcal{I}})\|_2}
        \;;\;
        \frac{f^{(\mathcal{C})}(q^{\mathcal{C}})}{\|f^{(\mathcal{C})}(q^{\mathcal{C}})\|_2}
      \,\right] \in \mathbb{R}^{n_{\mathcal{I}} + n_{\mathcal{C}}}
\end{equation}

\noindent\textbf{Retrieval Stage 1: Seed Node Retrieval.} We retrieve the top-$K$ product nodes most similar to $\mathbf{q}$ using a modality-weighted fused similarity, defined in Eqn. \ref{eq:seed-query-def} with fusion weight $\alpha \in [0,1]$. The result is set $\mathcal{S}$.

\begin{align}\label{eq:seed-query-def}
\begin{split}
    \mathcal{S} &= \operatorname*{arg\,top\text{-}K}_{v_i \in \mathcal{V}_I} \sigma(\mathbf{q}, \mathbf{h}_i^{P}), \\
    \sigma(\mathbf{q}, \mathbf{h}_i^{P}) &= \alpha\cdot \underbrace{\frac{{z_i^{\mathcal{I}\!\top}} f^{(\mathcal{I})}(q^{\mathcal{I}})}{\|z_i^{\mathcal{I}}\|_2\|f^{(\mathcal{I})}(q^{\mathcal{I}})\|_2}}_{\text{visual similarity}} \\
    &+ (1-\alpha)\cdot \underbrace{\frac{{\tilde{z}_i^{\mathcal{C}\!\top}} f^{(\mathcal{C})}(q^{\mathcal{C}})}{\|\tilde{z}_i^{\mathcal{C}}\|_2\|f^{(\mathcal{C})}(q^{\mathcal{C}})\|_2}}_{\text{attribute similarity}}
\end{split}
\end{align}

\noindent\textbf{Retrieval Stage 2: Hyperedge-Based Traversal.}
We collect all hyperedges incident to the seed set, $\mathcal{E}_{\mathcal{S}} = \{e \in \mathcal{E} \mid e \cap \mathcal{S} \neq \emptyset\}$, and score each by the mean node-type-conditioned relevance of its members as defined in Eqn. \ref{eq:node-sim}. 

\begin{equation}\label{eq:node-sim}
    s(\mathbf{q}, v) =
    \begin{cases}
        \sigma(\mathbf{q}, \mathbf{h}_{v}^{P}) & v \in \mathcal{V}_I \\[4pt]
        {\mathbf{q}^{\mathcal{D}}}^{\!\top} \mathbf{h}_{v}^{A} & v \in \mathcal{V}_A \\[4pt]
        {\mathbf{q}^{\mathcal{D}}}^{\!\top} \mathbf{h}_{v}^{T} & v \in \mathcal{V}_T \\[4pt]
        {\mathbf{q}^{\mathcal{D}}}^{\!\top} \mathbf{h}_{v}^{D} & v \in \mathcal{V}_P
    \end{cases}
\end{equation}

\begin{equation} \label{eq:hyperedge-reliability-score}
    \rho(e) = \frac{1}{|e|}\sum_{v \in e} s(\mathbf{q}, v)
\end{equation}

The top-$M$ hyperedges by $\rho(e)$ in Eqn. \ref{eq:hyperedge-reliability-score} form the query-dependent context $\mathcal{H}_q$. Based on experiments, we set $K = 8$ and $M = 10$ in our proposed configuration.

\noindent\textbf{Jurisdiction-Aware Context Assembly.}
The assembled context (Eqn. \ref{eq:proposed-context}) is drawn from two complementary paths to address the legislation semantics challenge. The \textit{hyperedge path} collects $\mathcal{C}_q^{\mathrm{hyp}} = \{p_v \mid v \in \mathcal{V}_q \cap \mathcal{V}_P\}$ and, for policy queries ($j_q \neq \varnothing$), surfaces the source chunks of the top-$T_{\mathrm{max}}$ entity nodes scored with a jurisdiction-aware boost as defined in Eqn. \ref{eq:vt-boost}.

\begin{equation}\label{eq:vt-boost}
    \tilde{s}_\phi(v_t) = s(\mathbf{q}, v_t) \cdot \beta^{\mathbf{1}[\mathrm{jur}(\phi(v_t)) = j_q]}, \;\; \beta = 2
\end{equation}

The ``boost factor'' $\beta = 2$ ensures that entity nodes whose source chunks originate from $j_q$ rank above semantically similar entities from other jurisdictions and non-legislative documents. The \textit{direct injection path} $\mathcal{C}_q^{\mathrm{jur}}$ scores all $\mathcal{V}_P$ nodes from $j_q$ directly against $\mathbf{q}^{\mathcal{D}}$ and injects the top-$P_{\mathrm{max}}$ not already present, activating unconditionally for all policy queries. In our experiments, we set $T_{\mathrm{max}} = 10$ and $P_{\mathrm{max}} = 5$.

\begin{equation}\label{eq:proposed-context}
    \mathcal{C}_q = \underbrace{\mathcal{C}_q^{\mathrm{hyp}} \;\cup\; \mathcal{C}_q^{\mathcal{V}_T}}_{\text{hyperedge path}} \;\cup\; \underbrace{\mathcal{C}_q^{\mathrm{jur}}}_{\text{direct injection}}
\end{equation}

\subsection{Response Generation} \label{subsec:response-gen}

With the assembled context $\mathcal{C}_q$ and user text query $q^{\mathcal{C}}$, the response from LLM $\pi$ is $y^{*} = \pi(q^{\mathcal{C}}, \mathcal{C}_q)$. Eqn. \eqref{eq:pipeline-formulation} provides a procedural summary of the query-time process for PRISM-RAG. 

\begin{equation} \label{eq:pipeline-formulation}
    (q^{\mathcal{I}}, q^{\mathcal{C}}) \xrightarrow{\text{\textbf{Stage 1}}\;} \mathcal{S} \xrightarrow{\text{\textbf{Stage 2}}\;} \mathcal{H}_q \xrightarrow{\text{assembly}\;} \mathcal{C}_q \xrightarrow{\pi} y^{*}
\end{equation}

\subsection{Cost Properties} \label{subsec:cost-theory}

We now cover cost properties that favor PRISM-RAG at index and query time compared to the SOTA RAG methods before moving on to empirical validation in Sec.~\ref{subsec:efficiency-analysis}. The SOTA RAG baselines, HippoRAG2 and HyperGraphRAG, both require one LLM call per document chunk to extract entities and relations at index time. For NicoPRISM, this amounts to 2,001 API calls for our knowledge base. Meanwhile, PRISM-RAG's spaCy-based entity extraction (Sec.~\ref{subsec:entity-extraction}) requires no LLM calls. At query time, both PRISM-RAG and StandardRAG each require exactly one LLM call, versus two for HyperGraphRAG (one internal to its knowledge-graph traversal, then another for response generation).

PRISM-RAG's direct injection path also provides a structural guarantee on jurisdiction routing, formalized below. Let $q$ be a policy query with jurisdiction hint $j_q \neq \varnothing$, let $P_{\max}=5$ be the fixed direct-injection budget (Sec.~\ref{subsec:mm-query-procedure}), and let $n_{j_q} = |\{v \in \mathcal{V}_P : \mathrm{jur}(v) = j_q\}|$ be the number of document chunks in the knowledge base with jurisdiction label $j_q$. The context $\mathcal{C}_q$ assembled by PRISM-RAG (Eqn.~\ref{eq:proposed-context}) satisfies the property given in Eqn. \eqref{eq:ja-guarantee}, completely independent of knowledge organization (graph or hypergraph) or embedding-space topology.

\begin{equation} \label{eq:ja-guarantee}
    |\{v \in \mathcal{C}_q : \mathrm{jur}(v) = j_q\}| \;\geq\; \min(P_{\max},\, n_{j_q}),
\end{equation}

The direct injection path $\mathcal{C}_q^{\mathrm{jur}} \subseteq \mathcal{C}_q$ scores every $v \in \mathcal{V}_P$ with $\mathrm{jur}(v)=j_q$ against the query embedding and selects the top-$P_{\max}$ unconditionally whenever $j_q \neq \varnothing$, independent of Stage 1 or Stage 2 outcomes. If $n_{j_q} \geq P_{\max}$, exactly $P_{\max}$ chunks are injected; otherwise all $n_{j_q}$ available chunks are injected. In both cases, the bound is satisfied.

The bound is only meaningful when the queried jurisdiction is represented in the knowledge base ($n_{j_q} > 0$); if $j_q$ is entirely absent from NicoPRISM's document corpus, the guarantee holds trivially ($\geq 0$) but provides no coverage. This case, however, we note, is out of the scope of PRISM-RAG. We also highlight the importance of incorporating external knowledge in future work.

\section{Experiments} \label{sec:experiments}

We describe our experiment evaluation setup in Sec. \ref{subsec:experimental-setup}. Sec. \ref{subsec:main-qa} reports primary benchmark results on policy compliance QA, the primary task of this work. Sec. \ref{subsec:efficiency-analysis} provides a cost analysis between PRISM-RAG and the other RAG methods. Sec. \ref{subsec:ablations} presents ablation studies alongside product knowledge QA results. Sec. \ref{subsec:product-retrieval} evaluates image encoder backbones on visual product retrieval. Sec. \ref{subsec:qualitative} provides a qualitative analysis of RAG framework responses.

\subsection{Experimental Setup} \label{subsec:experimental-setup}

For image embeddings and all cross-modal operations, we use SigLIP \cite{zhai2023sigmoidlosslanguageimage} (SoViT-400M, patch size 14, input size 384); for embedding longer documents locally, as well as text within PRISM-RAG, we use BGE-M3 \cite{bge-m3_2024}. Our LLM of choice for all RAG backends and final response generation is GPT-5.4-Nano \cite{singh2025openaigpt5card}. For QA evaluation, we use a held-out test set of 275 images, and the test image-QA pairings give 1,325 policy compliance instances and 170 product knowledge instances. Visual product retrieval results are averaged over five random 70-30 splits $\{0, 42, 101, 777, 1000\}$.

We evaluate three document RAG baselines alongside PRISM-RAG: \textbf{StandardRAG} (document chunk retrieval); \textbf{HippoRAG2} \cite{jimenez2024hipporag} (graph-based LLM entity extraction at index time); and \textbf{HyperGraphRAG} \cite{luo2025hypergraphrag} ($n$-ary knowledge hyperedges with LLM calls at index time). All baselines use $K=8$ retrieved products and top-8 retrieved documents. All PRISM-RAG experiments use bimodal product node embedding with SigLIP (image) and BGE-M3 (text) branches, mutual top-$k$ bridge matching with $k=10$, $k$-means concept clustering with $c=50$ clusters, entity extraction using spaCy, $K=8$ seed nodes at Stage 1, and a traversal depth of $m=10$ hyperedges at retrieval Stage 2. These hyperparameters were selected by maximizing JA on the policy compliance task as reported in Table \ref{tab:policy_qa_metrics_ablation}. 

The metrics are defined as follows. We report keyword precision (\textbf{KW-P}), keyword recall (\textbf{KW-R}), and F1 (\textbf{KW-F1}) scores to quantify retrieval success of ground-truth keywords/entities in $y^*$, as is the case in past RAG works. Furthermore, we introduce context coverage (\textbf{CC})---the fraction of ground-truth keywords present anywhere in $\mathcal{C}_q$, \ie, measuring \textit{retrieval quality} independently of generation quality. Response groundedness \textbf{RG} (response groundedness) measures similarity between $y^*$ and $\mathcal{C}_q$ via cosine similarity. We also employ LLM-as-a-Judge (\textbf{Judge}) score (range 1--5) \cite{que2024hellobench}. 
For policy compliance questions, compliance accuracy (\textbf{CA}) is the fraction of responses whose LLM-judge-extracted compliance label matches the ground-truth label; and jurisdiction accuracy (\textbf{JA}) is the fraction of retrieved passages whose source document belongs to the jurisdiction queried by the user, denoted as $j_q$, and is either a single jurisdiction for most queries but a set of two jurisdictions for multi-hop preemption questions (\eg, ``legal in State A but banned in State B''); a passage counts as correct-jurisdiction if its source jurisdiction is a member of $j_q$, computed from an external document-to-jurisdiction mapping.
Lastly, bootstrap 95\% confidence intervals and BH-corrected ($q = 0.05$) paired permutation test $p$-values over all 24 simultaneously tested hypotheses are reported in the Appendix.

We note that we do not compare against VLM-based RAG methods (\eg, multimodal retrieval augmented generation with GPT-4V or similar) because the rate-limiting step in NicoPRISM is not visual recognition but legislative document retrieval---distinguishing city (\eg, D.C.) from federal statutory text, for example, requires \textit{document-level structural understanding} that VLM context windows and short-form image-text alignment are not designed to provide. The selected baselines represent the current state of the art specifically for document-grounded knowledge graph and hyperedge retrieval, which is the capability most directly relevant to this task.

\subsection{Policy Compliance QA} \label{subsec:main-qa}


\begin{table*}[!t]
\centering
\caption{
    Policy compliance QA results on 1,325 test instances spanning 13 jurisdictions.
    \textbf{KW-R} = keyword recall; \textbf{KW-F1} = keyword F1;
    \textbf{CC} = context coverage; \textbf{JA} = retrieval-level jurisdiction accuracy;
    \textbf{CA} = compliance accuracy; \textbf{Judge} = LLM-as-judge score (1--5);
    \textbf{RG} = response groundedness.
    All PRISM-RAG results use $m=10$, $c=50$.
    $\dagger$ HyperGraphRAG makes two LLM calls and three API calls total per query.
    Bootstrap 95\% CIs and BH-corrected $p$-values for all pairwise comparisons are reported in the Appendix.
}
\label{tab:policy_qa_metrics_main}
\begin{tabular}{lrrrrrrr}
\toprule
\textbf{Method} & \textbf{KW-R} & \textbf{KW-F1} & \textbf{CC} & \textbf{JA} & \textbf{CA} & \textbf{Judge} & \textbf{RG} \\
\midrule
StandardRAG             & 0.2683 & 0.1821 & 0.3942 & 0.4495 & 0.3911 & 2.7442 & 0.7336 \\
HippoRAG2               & 0.0507 & 0.0345 & 0.5778 & 0.1408 & 0.0844 & 1.4106 & 0.4725 \\
HyperGraphRAG$^\dagger$ & 0.2220 & 0.1475 & 0.2816 & 0.8773 & 0.0133 & 2.7283 & \textbf{0.8850} \\
\midrule
\textbf{PRISM-RAG (ours)} & \textbf{0.3139} & \textbf{0.1997} & \textbf{0.5495} & \textbf{0.9387} & \textbf{0.4178} & \textbf{2.8551} & 0.7287 \\
\bottomrule
\end{tabular}
\end{table*}

\begin{table*}[!t]
\centering
\small
\caption{
    Pairwise mean gaps (PRISM-RAG minus baseline) with bootstrap 95\% confidence intervals and BH-corrected $p$-values on policy compliance QA. The star $^*$ means a significant result at adjusted $q < 0.05$ (BH correction over all 24 simultaneously tested hypotheses across metrics and baseline pairs; 10,000 bootstrap resamples and 10,000 permutations). CA uses $N = 235$ matched instances with an extractable compliance label; all other metrics use $N = 1{,}387$ matched pairs ($N = 1{,}231$ for PRISM-RAG vs.\ HyperGraphRAG on JA). 
}
\label{tab:policy_qa_ci}
\resizebox{\textwidth}{!}{
\begin{tabular}{lcccc}\toprule
\textbf{Comparison}
    & \textbf{JA} [95\% CI]
    & \textbf{CA} [95\% CI]
    & \textbf{KW-F1} [95\% CI]
    & \textbf{Judge} [95\% CI] \\\midrule
vs.\ StandardRAG
    & $+0.486\ [0.462,\ 0.509]^*$
    & $+0.030\ [{-0.034},\ 0.094]\ $
    & $+0.019\ [0.014,\ 0.023]^*$
    & $+0.114\ [0.043,\ 0.180]^*$ \\
vs.\ HippoRAG2
    & $+0.796\ [0.781,\ 0.811]^*$
    & $+0.340\ [0.281,\ 0.404]^*$
    & $+0.166\ [0.159,\ 0.173]^*$
    & $+1.455\ [1.381,\ 1.528]^*$ \\
vs.\ HyperGraphRAG
    & $+0.061\ [0.041,\ 0.081]^*$
    & $+0.409\ [0.345,\ 0.472]^*$
    & $+0.053\ [0.049,\ 0.058]^*$
    & $+0.136\ [0.049,\ 0.219]^*$ \\\bottomrule
\end{tabular}
}
\end{table*}

Table \ref{tab:policy_qa_metrics_main} reports results on policy compliance QA.

\noindent\textbf{Jurisdiction Accuracy} is our most important finding. PRISM-RAG retrieves passages from the queried jurisdiction in 93.9\% of instances versus 45.0\% for StandardRAG, and this +48.6~pp advantage is statistically unambiguous ($p < 0.001$, bootstrap 95\% CI: [46.2, 50.9]). Meanwhile, HyperGraphRAG achieves JA of 87.7\% and is within 6.1~pp of PRISM-RAG's own JA, therefore suggesting its knowledge graph (KG) traversal reaches jurisdiction-relevant content relatively often. Performance on CA, however, collapses to 1.3\% and CC collapses to 0.2816, the lowest among the RAG methods tested. Despite this discrepancy, HyperGraphRAG's responses are highly grounded in the retrieved context (RG = 0.885, highest of all methods) but do not produce actionable compliance determinations. This ``grounded but indeterminate'' pattern reflects the fact that HyperGraphRAG's entity-relation KG surfaces topically adjacent content without producing the specific banned/restricted/legal conclusion the judge extracts for CA computation, even though it does retrieve jurisdictionally relevant documents as its JA score shows. We call this phenomenon the ``JA-CA dissociation.'' Interestingly, this JA-CA dissociation pattern from the main results occurs again across the ablation experiments in Table \ref{tab:policy_qa_metrics_ablation}, where several PRISM-RAG variants that reduce JA simultaneously raise CA. We argue this suggests that neither metric alone is sufficient and that \textit{both} JA and CA measure complementary properties of the retrieval method's pipeline. We should also recall that HyperGraphRAG requires two LLM calls per query (one for entity extraction, one for final response generation) as opposed to one per query for PRISM-RAG with no API costs for indexing.

\noindent\textbf{Compliance accuracy} reaches 41.8\% for PRISM-RAG versus 39.1\% for StandardRAG. The CA gap is not statistically significant ($p = 0.430$, bootstrap 95\% CI on gap: [$-$0.034, 0.094], $N = 235$ matched instances with extractable compliance labels). This shows why multimodal legislation understanding is challenging. PRISM-RAG reaches the correct jurisdiction more than 2.1$\times$ more often than StandardRAG, as shown by JA. Thus, while the LLM can sometimes draw the right compliance conclusion, it does so because it \textit{pulled information from the wrong jurisdiction}, because statutory language is sufficiently similar across jurisdictions. Additionally, HippoRAG2 achieves CA of 8.4\% and HyperGraphRAG 1.3\%, showing their overreliance on semantics to draw out text entities. All KW-R, KW-F1, CC, and Judge gaps between PRISM-RAG and HippoRAG2 or HyperGraphRAG are statistically significant at $p < 0.001$ after BH correction; full CI tables are in the Appendix.

\noindent\textbf{Context Coverage} further illuminates retrieval behavior in the SOTA RAG frameworks. HippoRAG2 achieves CC of 0.578 (highest among baselines) because its KG traversal surfaces a large, noisy context that by chance includes many keyword strings, yet its KW-R is only 0.051---the LLM cannot extract or reproduce the relevant information from the overly long context. Among the RAG methods used, HippoRAG2 has the longest contexts and consumes the most tokens. PRISM-RAG achieves CC of 0.550 alongside KW-R of 0.314 and a 4.1$\times$ wall-time advantage, suggesting that smaller and more precise, better-ordered contexts produced by jurisdiction-aware assembly are more effective.

\noindent\textbf{Confidence Intervals and Significance Tests on Policy QA}. Table \ref{tab:policy_qa_ci} reports bootstrap 95\% confidence intervals (CIs) and Benjamini-Hochberg (BH)-corrected pairwise significance tests for the key policy compliance metrics. To ensure that our improvements in high-similarity document chunk retrieval in our RAG experiments are statistically sound and not artifacts of multiple testing rounds, we do not use the standard Student's $t$-test and instead use a BH-corrected paired permutation test. The JA gap between PRISM-RAG and StandardRAG is the largest and most precisely estimated result in the table at +48.6~pp [46.2,\;50.9], $p < 0.001$. The CA gap against StandardRAG is +3.0~pp [$-$3.4,\;9.4], $p = 0.430$, which is not statistically significant after BH correction. The wide CI for CA reflects that only $N = 235$ of the 1,325 instances yielded a clear compliance label in both RAG backends' outputs, leaving the comparison underpowered relative to the other metrics. CA parity with StandardRAG does not imply equivalent retrieval behavior; JA reveals the underlying difference directly. For instance, PRISM-RAG reaches the correct jurisdiction 2.1$\times$ more often (JA gap: +48.6~pp), however, StandardRAG can still produce a correct compliance label when the wrong jurisdiction's statute happens to share the same regulatory stance as the queried one. All gaps against HippoRAG2 and HyperGraphRAG are statistically significant at $p < 0.001$ across JA, CA, KW-F1, and Judge. The sole exception is on RG, where HyperGraphRAG's responses more closely mirror their retrieved context ($-$0.157 [$-$0.162,\;$-$0.152], $p < 0.001$), consistent with the grounded-but-indeterminate pattern discussed above. Full bootstrap CIs per method and across all metrics appear in the Appendix. 

\subsection{Efficiency Analysis} \label{subsec:efficiency-analysis}

The cost properties discussion in Sec. \ref{subsec:cost-theory} established that PRISM-RAG and StandardRAG each require exactly one LLM call to process one query, HyperGraphRAG requires two, and HippoRAG2 and HyperGraphRAG additionally require one LLM call per document chunk at index time, while PRISM-RAG requires none. Table \ref{tab:efficiency} reports the measured wall-time and token cost consequences of these structural differences, using the same 1,495-query benchmark and \texttt{gpt-5.4-nano} backbone as Sec. \ref{subsec:main-qa}.

\begin{table*}[!ht]
\centering\small
\caption{
    Index-time and query-time cost across RAG methods.
    \textbf{Index LLM} = LLM calls at index time; \textbf{Query LLM} = LLM calls per query;
    \textbf{WT} = mean wall time per query (s/q); \textbf{Avg.\ Tok} = mean input tokens per LLM call with respect to \texttt{gpt-5.4-nano}.
    $|\mathcal{D}|=2{,}001$ document chunks from NicoPRISM's document corpus.
}
\label{tab:efficiency}
\begin{tabular}{lcccc}\toprule
\textbf{Method} & \textbf{Index LLM} & \textbf{Query LLM} & \textbf{WT (s/q)} & \textbf{Avg.\ Tok} \\\midrule
StandardRAG & 0 & 1 & 1.99 & 3970 \\
HippoRAG2 & $|\mathcal{D}|$ & 1 & 24.15 & 366310 \\
HyperGraphRAG & $|\mathcal{D}|$ & 2 & 10.88 & 1438 \\
\textbf{PRISM-RAG (ours)} & \textbf{0} & \textbf{1} & \textbf{3.37} & \textbf{4331} \\
\bottomrule
\end{tabular}
\end{table*}

The wall-time ordering is broadly consistent with the call-count differences established above: HyperGraphRAG runs approximately 3.2$\times$ slower per query than PRISM-RAG (10.88~s vs.\ 3.37~s). StandardRAG is naturally the fastest method overall (1.99~s/q), 1.7$\times$ faster than PRISM-RAG. This speed advantage, however, comes at a substantial accuracy cost; recall that StandardRAG's JA trails PRISM-RAG's by 48.6~pp with $p<0.001$ (refer to Table \ref{tab:policy_qa_ci}, Sec. \ref{subsec:main-qa}). The tradeoff for PRISM-RAG's modest added latency affords it a substantial, and statistically robust, jurisdiction-routing advantage that StandardRAG cannot offer at any query-time cost. Furthermore, HippoRAG2 is the slowest method (24.15~s/q, 7.2$\times$ PRISM-RAG) and consumes 366{,}310 input tokens per query on average, or nearly 85$\times$ PRISM-RAG's average of 4{,}331 tokens per query. This finding is consistent with the unbounded and noisy retrieval behavior identified via the CC-KW-R dissociation in Sec. \ref{subsec:main-qa} and the query-time context-length failures in our experiments.

\subsection{Ablation Studies} \label{subsec:ablations}

\begin{table*}[!t]
\centering
\caption{
    Policy compliance QA ablation results.
    \textbf{CC} = context coverage; \textbf{JA} = retrieval-level jurisdiction accuracy;
    \textbf{CA} = compliance accuracy; \textbf{Judge} = LLM-as-judge score (1--5).
    JA is the primary ablation metric; CA is reported alongside it but exhibits a JA-CA dissociation in several configurations (see text). Unless otherwise noted, all ablations use $c=50$, $m=10$.
}
\label{tab:policy_qa_metrics_ablation}
\begin{tabular}{lrrrrrrr}
\toprule
\textbf{Method} & \textbf{KW-P} & \textbf{KW-R} & \textbf{KW-F1} & \textbf{CC} & \textbf{JA} & \textbf{CA} & \textbf{Judge} \\
\midrule
\multicolumn{8}{l}{\textit{Index-time: bridge method and entity extractor}} \\
PRISM-RAG (ours)             & 0.1626 & 0.3139 & 0.1997 & 0.5495 & \textbf{0.9387} & \textbf{0.4178} & 2.8551 \\
PRISM-RAG$_\text{bridge}$    & 0.1642 & 0.3140 & 0.2008 & 0.5565 & 0.9396          & 0.4000          & 2.8891 \\
PRISM-RAG$_\text{LLM}$       & 0.1611 & 0.3147 & 0.1985 & 0.5629 & 0.8946          & 0.4489          & 2.8415 \\
PRISM-RAG$_\text{ontology}$  & 0.1639 & 0.3125 & 0.2000 & 0.5548 & 0.9094          & 0.3778          & 2.8838 \\
PRISM-RAG$_\text{agglo}$     & 0.1618 & 0.3061 & 0.1970 & 0.5342 & 0.8682          & 0.4356          & 2.8581 \\
\midrule
\multicolumn{8}{l}{\textit{Index-time: cluster count ($m=10$)}} \\
PRISM-RAG$^{c=50}$ (ours) & 0.1626 & 0.3139 & 0.1997 & 0.5495 & \textbf{0.9387} & \textbf{0.4178} & 2.8551 \\
PRISM-RAG$^{c=100}$       & 0.1693 & 0.3201 & 0.2059 & 0.5451 & 0.9170          & 0.3556          & 2.9200 \\
PRISM-RAG$^{c=150}$       & 0.1627 & 0.3086 & 0.1984 & 0.5382 & 0.8439          & 0.4222          & 2.8438 \\
PRISM-RAG$^{c=200}$       & 0.1588 & 0.3058 & 0.1942 & 0.5422 & 0.8255          & 0.4267          & 2.8279 \\
\midrule
\multicolumn{8}{l}{\textit{Query-time: traversal depth $m$ ($c=50$)}} \\
PRISM-RAG$_{m=3}$          & 0.1603 & 0.3024 & 0.1946 & 0.5293 & 0.8702          & 0.4044          & 2.8121 \\
PRISM-RAG$_{m=5}$          & 0.1628 & 0.3091 & 0.1985 & 0.5480 & 0.8876          & 0.3689          & 2.8574 \\
PRISM-RAG$_{m=8}$          & 0.1649 & 0.3136 & 0.2013 & 0.5585 & 0.9178          & 0.3467                              & 2.8785 \\
PRISM-RAG$_{m=10}$ (ours)  & 0.1626 & 0.3139 & 0.1997 & 0.5495 & \textbf{0.9387} & \textbf{0.4178} & 2.8551 \\
PRISM-RAG$_{m=15}$         & 0.1677 & 0.3148 & 0.2043 & 0.5596 & 0.9251          & 0.4044          & 2.9064 \\
\midrule
\multicolumn{8}{l}{\textit{Query-time: traversal depth $m$ ($c=100$)}} \\
PRISM-RAG$_{m=3}$   & 0.1582 & 0.3029 & 0.1934 & 0.5370 & 0.7020 & 0.4267 & 2.8030 \\
PRISM-RAG$_{m=5}$   & 0.1613 & 0.3050 & 0.1965 & 0.5415 & 0.7540 & 0.4356 & 2.8151 \\
PRISM-RAG$_{m=8}$   & 0.1626 & 0.3103 & 0.1986 & 0.5472 & 0.8187 & 0.4178                     & 2.8551 \\
PRISM-RAG$_{m=10}$  & 0.1597 & 0.3042 & 0.1952 & 0.5451 & 0.8549 & 0.3600 & 2.8309 \\
PRISM-RAG$_{m=15}$  & 0.1624 & 0.3085 & 0.1982 & 0.5444 & 0.8887 & 0.4222                     & 2.8513 \\
\bottomrule
\end{tabular}
\end{table*}

\begin{table}[!ht]
\centering
\caption{
    Product knowledge QA ablation results.
    \textbf{CC} = context coverage.
    Performance is broadly uniform across ablations, consistent with this task being bounded by knowledge-base coverage rather than retrieval architecture. Unless otherwise noted, all ablations use $c=50$, $m=10$.
}
\label{tab:product_qa_metrics_ablation}
\begin{tabular}{lrrrr}
\toprule
\textbf{Method} & \textbf{KW-P} & \textbf{KW-R} & \textbf{KW-F1} & \textbf{CC} \\
\midrule
\multicolumn{5}{l}{\textit{Main comparison (from Table~\ref{tab:policy_qa_metrics_main})}} \\
StandardRAG         & 0.1518 & 0.1935 & 0.1360 & 0.1520 \\
HippoRAG2           & 0.0326 & 0.0137 & 0.0173 & 0.3395 \\
HyperGraphRAG       & 0.1189 & 0.1577 & 0.1069 & 0.0473 \\
PRISM-RAG (ours)    & 0.1322 & 0.2027 & 0.1342 & 0.1552 \\
\midrule
\multicolumn{5}{l}{\textit{Index-time: bridge method and entity extractor}} \\
PRISM-RAG (ours)             & 0.1322 & 0.2027 & 0.1342 & 0.1552 \\
PRISM-RAG$_\text{bridge}$    & 0.1351 & 0.1962 & 0.1331 & 0.1785 \\
PRISM-RAG$_\text{LLM}$       & 0.1376 & 0.1850 & 0.1296 & 0.1221 \\
PRISM-RAG$_\text{ontology}$  & 0.1424 & 0.1941 & 0.1355 & 0.1714 \\
PRISM-RAG$_\text{agglo}$     & 0.1534 & 0.2020 & 0.1353 & 0.0533 \\
\midrule
\multicolumn{5}{l}{\textit{Index-time: cluster count ($m=10$)}} \\
PRISM-RAG$^{c=50}$ (ours) & 0.1322 & 0.2027 & 0.1342 & 0.1552 \\
PRISM-RAG$^{c=100}$       & 0.1401 & 0.1987 & 0.1326 & 0.1759 \\
PRISM-RAG$^{c=150}$       & 0.1424 & 0.1975 & 0.1319 & 0.1520 \\
PRISM-RAG$^{c=200}$       & 0.1390 & 0.2063 & 0.1365 & 0.0942 \\
\midrule
\multicolumn{5}{l}{\textit{Query-time: traversal depth $m$ ($c=50$)}} \\
PRISM-RAG$_{m=3}$          & 0.1439 & 0.2044 & 0.1359 & 0.1159 \\
PRISM-RAG$_{m=5}$          & 0.1389 & 0.1962 & 0.1333 & 0.1435 \\
PRISM-RAG$_{m=8}$          & 0.1432 & 0.1977 & 0.1356 & 0.1554 \\
PRISM-RAG$_{m=10}$ (ours)  & 0.1322 & 0.2027 & 0.1342 & 0.1552 \\
PRISM-RAG$_{m=15}$         & 0.1477 & 0.1959 & 0.1421 & 0.1695 \\
\midrule
\multicolumn{5}{l}{\textit{Query-time: traversal depth $m$ ($c=100$)}} \\
PRISM-RAG$_{m=3}$   & 0.1417 & 0.1980 & 0.1349 & 0.0593 \\
PRISM-RAG$_{m=5}$   & 0.1400 & 0.1932 & 0.1305 & 0.0885 \\
PRISM-RAG$_{m=8}$   & 0.1384 & 0.2022 & 0.1334 & 0.1003 \\
PRISM-RAG$_{m=10}$  & 0.1295 & 0.1888 & 0.1275 & 0.1158 \\
PRISM-RAG$_{m=15}$  & 0.1360 & 0.1860 & 0.1286 & 0.1530 \\
\bottomrule
\end{tabular}
\end{table}

Tables \ref{tab:product_qa_metrics_ablation} and \ref{tab:policy_qa_metrics_ablation} report ablation results for PRISM-RAG across both QA tasks. Moreover, Table \ref{tab:product_qa_metrics_ablation} serves as the primary results table for the product knowledge QA task, where performance is broadly uniform across all ablations. This is consistent with knowledge on products depending on a breadth of knowledge rather than retrieval architecture. Further discussion is included in the Appendix. We focus the ablation analysis on policy compliance QA (Table \ref{tab:policy_qa_metrics_ablation}), where maximizing JA is our primary goal.

\noindent\textbf{Preliminary: the JA-CA dissociation.}
Several PRISM-RAG ablation configurations exhibit lower JA but higher CA than the proposed setting. This occurs because CA is a downstream metric that conflates retrieval correctness with jurisdictional regulatory alignment. That is, a RAG method can produce the correct compliance label despite it retrieving the wrong jurisdiction's statute, provided both jurisdictions happen to share similar regulatory stance (\eg, retrieving Massachusetts content for a California compliance query may still yield ``banned'' because both states ban flavored nicotine pouches). Therefore, we chose JA is the cleaner measure of structural retrieval correctness, and treat it as the primary ablation criterion accordingly. The proposed configuration ($c=50$, $m=10$) significantly exceeds every evaluated configuration on JA except PRISM-RAG$_\text{bridge}$, against which the gap is not significant ($p=0.241$); no configuration's CA differs significantly from the proposed setting's (all CA gaps $p>0.05$).

\noindent\textbf{Entity extraction.} LLM-powered entity extraction in PRISM-RAG reduces JA by 4.4~pp (0.895 vs.\ 0.939, $p<0.001$) while its CA point estimate is 3.1~pp higher (but not statistically significant, $p=0.611$). LLM-extracted entities tend to be relational abstractions (\eg, ``California prohibits flavored tobacco products'') rather than the straightforward noun phrases spaCy extracts (\eg, ``flavored tobacco product'', ``characterizing flavor''). These abstractions embed differently from product attribute nodes and cluster less naturally with them in concept hyperedge construction, weakening the traversal path into jurisdiction-specific legislative content. Spacy extraction is therefore preferred with the added benefit of zero index-time LLM cost.

\noindent\textbf{Ontology augmentation.} Augmenting legislative entity text with domain synonyms, meant to make PRISM-RAG's indexing more flexible, before encoding degrades JA by 2.9~pp ($p<0.001$); its CA point estimate is lower but not significantly ($p=0.281$).

\noindent\textbf{Agglomerative clustering.} Replacing $k$-means with agglomerative clustering reduces JA by 7.1~pp (0.868 vs.\ 0.939, $p<0.001$), while its CA point estimate is approximately 1.7~pp higher (not significant, $p=0.755$).

\noindent\textbf{Cluster count $c$.} 
JA decreases monotonically as $c$ increases: 0.939 ($c=50$) $>$ 0.917 ($c=100$) $>$ 0.844 ($c=150$) $>$ 0.826 ($c=200$), all significant drops relative to $c=50$ ($p<0.001$ in each case), while CA differences from $c=50$ are never significant across this sweep. Finer-grained clusters produce smaller and more semantically specific hyperedges, reducing the probability that product attribute nodes and legislative entity nodes are co-clustered. This mechanism by which concept hyperedges bridge marketing language to statutory terminology is critical. Thus, $c=50$ is optimal for both JA and CA.

\noindent\textbf{Traversal depth $m$ ($c=50$).} JA increases from $m=3$ (0.870) to $m=10$ (0.939) then dips at $m=15$ (0.925), making $m=10$ the optimal traversal depth; every $m \neq 10$ value in this sweep differs from $m=10$ at $p<0.001$ on JA. The drop at $m=15$ is consistent with context dilution, where at high traversal depths, the sub-hypergraph $\mathcal{H}_q$ includes off-jurisdiction hyperedges which introduce noise. CA point estimates follow the opposite pattern (decreases as $m$ increases, but highest at $m=10$, and then decreases), though none of the CA differences from $m=10$ in this sweep are statistically significant.

\noindent\textbf{Traversal depth $m$ ($c=100$).} With $c=100$, JA increases monotonically across all evaluated $m$ values, from 0.702 at $m=3$ up to 0.889 at $m=15$), however, no $c=100$ configuration reaches the JA of the proposed $c=50$, $m=10$ setting. The highest $c=100$ result ($m=15$, JA~$=$~0.889) is 5.0~pp below the proposed configuration ($p<0.001$), confirming that the coarser cluster granularity of $c=50$ is the primary driver of jurisdiction-routing accuracy rather than the traversal depth. Notably, PRISM-RAG$_{m=3,c=100}$ is the lowest scoring JA configuration overall (0.702), and significantly \textit{exceeds} the proposed configuration's KW-F1 ($p<0.001$), showing that reliance on keyword overlap can be a misleading sign for jurisdictional correctness.

\subsection{Visual Product Retrieval} \label{subsec:product-retrieval}

We benchmark six image encoder backbones on brand-level product retrieval across all three NicoPRISM image sets using five random 70-30 knowledge-query splits. Full precision and accuracy tables appear in the Appendix. 
SigLIP ($14\times14$, WebLi) leads on within-set and cross-set retrieval across both web and TikTok sets. We note that cross-set results quantify the image distribution gap between the web set and social media frames, which we identify as a key direction for future work.

\subsection{Qualitative Analysis} \label{subsec:qualitative}

%
%
\begin{figure*}[!ht]
\centering
\footnotesize
\newlength{\qboxwidth}
\setlength{\qboxwidth}{\textwidth}
\begin{tcolorbox}[
    enhanced,
    width=\qboxwidth,
    colback=gray!7,
    colframe=gray!45,
    fonttitle=\large\bfseries,
    title={Query},
    boxrule=0.50pt,
    top=4pt, bottom=4pt, left=8pt, right=8pt,
    sidebyside, sidebyside align=top,
    lefthand width=0.12\textwidth,
    segmentation style={solid, gray!40, line width=0.4pt},
]
\centering
\includegraphics[width=\linewidth,height=6cm,keepaspectratio]{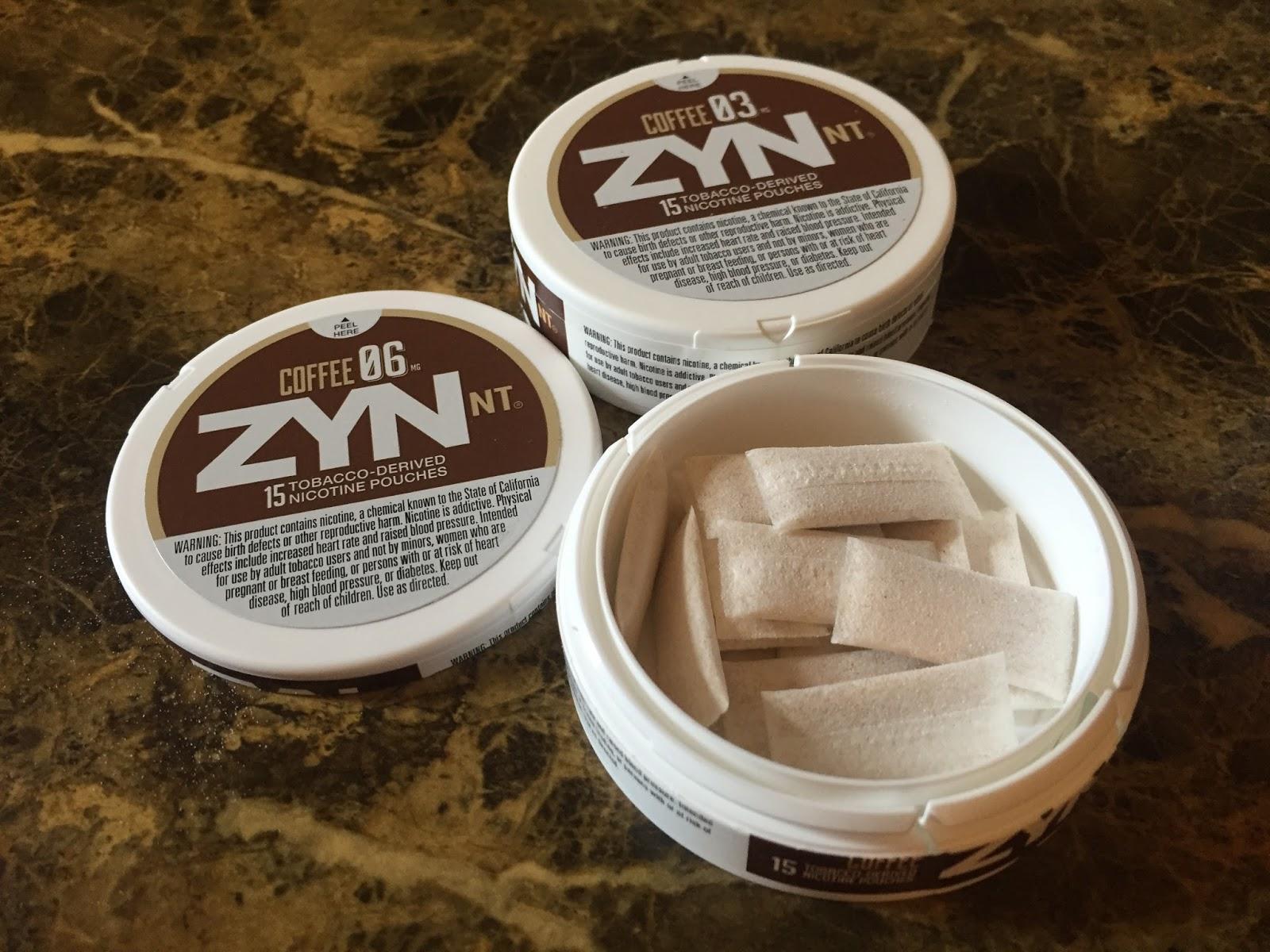}
\tcblower
\large
Are nicotine pouches regulated as tobacco products under Washington,~D.C.'s statutes,
or do they fall under a separate nicotine product framework?
\end{tcolorbox}

\vspace{2pt}

\begin{tcolorbox}[
    enhanced,
    width=\qboxwidth,
    colback=blue!4,
    colframe=blue!38,
    fonttitle=\normalsize\bfseries,
    title={PRISM-RAG\hfill
           \normalfont\normalsize
           KW-R:\,\textbf{0.40}~~
           KW-P:\,\textbf{0.261}~~
           KW-F1:\,\textbf{0.316}~~
           Judge:\,\textbf{5.0\,/\,5}~~
           JA:\,\textbf{1.0}},
    boxrule=0.70pt,
    top=4pt, bottom=4pt, left=6pt, right=6pt,
]
\small
Based on the DC statutes described in the provided knowledge-base excerpts,
nicotine pouches are treated under
\colorbox{green!18}{\strut\textbf{the District's ``tobacco products'' framework}},
not a separate nicotine-only framework.
\smallskip
\begin{itemize}[leftmargin=1.2em, itemsep=2pt, topsep=2pt]
  \item District law's
        \colorbox{green!18}{\strut flavored-tobacco ban}
        applies to \colorbox{green!18}{\strut\textbf{``tobacco products''}}
        and defines tobacco products as including
        \colorbox{green!18}{\strut\textbf{``consumable tobacco or synthesized nicotine products.''}}
        (\emph{Knowledge base excerpt summarising}
         \colorbox{orange!20}{\strut D.C. Official Code \S\,7-1721.01(1) / \S\,7-1721.08}.)
  \item The flavored ban also explicitly references
        \colorbox{green!18}{\strut\textbf{``flavored nicotine or synthetic nicotine products''}}
        as covered flavored tobacco products.
        (\emph{Knowledge base excerpt for}
         \colorbox{orange!20}{\strut D.C. Official Code \S\,7-1721.08(b)}.)
\end{itemize}
\smallskip
So, \textbf{nicotine pouches (including flavored ones)} would fall within
DC's \textbf{tobacco-products} regime for purposes of the
\colorbox{green!18}{\strut flavored-tobacco restrictions},
rather than being regulated under a separate nicotine-product-only framework
(at least as reflected in these excerpts).
\end{tcolorbox}
\vspace{2pt}
\begin{tcolorbox}[
    enhanced,
    width=\qboxwidth,
    colback=orange!4,
    colframe=orange!45,
    fonttitle=\normalsize\bfseries,
    title={StandardRAG\hfill
           \normalfont\normalsize
           KW-R:\,\textbf{0.20}~~
           KW-P:\,\textbf{0.097}~~
           KW-F1:\,\textbf{0.131}~~
           Judge:\,\textbf{2.0\,/\,5}~~
           JA:\,\textbf{0.0}},
    boxrule=0.70pt,
    top=4pt, bottom=4pt, left=6pt, right=6pt,
]
\small
Based on the retrieved records/knowledge excerpts, nicotine pouches are treated
as \colorbox{green!18}{\strut\textbf{tobacco products}}
under \textbf{federal FDA oversight}
(\ie, they fall under the Family Smoking Prevention and Tobacco Control Act
framework), even though they typically \textbf{do not contain tobacco leaf}---they
contain nicotine that is \textbf{tobacco-derived or synthesized}.
\smallskip
However, the question asks specifically about \textbf{Washington,~DC statutes}.
The provided materials \textbf{do not include DC-specific statutory text or a DC
regulatory classification} for nicotine pouches versus a separate
``nicotine product'' framework. So, from the records/excerpts available here,
I\,\textbf{cannot confirm} whether Washington,~DC:
\begin{itemize}[leftmargin=1.2em, itemsep=2pt, topsep=2pt]
  \item classifies nicotine pouches as
        \colorbox{green!18}{\strut\textbf{``tobacco products'' under DC law}},~or
  \item instead regulates them under a \textbf{separate nicotine-product scheme}.
\end{itemize}
If you can share the relevant DC code sections or an excerpt
(\eg, the definitions of ``tobacco product'' and/or ``nicotine product''),
I can apply them directly to these products (\eg, ZYN/ZYN~NT nicotine pouches).
\end{tcolorbox}
\caption{
    Qualitative comparison of PRISM-RAG and StandardRAG on a Washington,~D.C.\ policy compliance question.
    \colorbox{green!18}{Green highlights} mark ground-truth keywords that appear in the response (contributing to KW-R); \colorbox{orange!20}{orange highlights} mark DC statutory citations (\S\,7-1721.01(1), \S\,7-1721.08(b)) that are correct but excluded from KW-R computation due to the `\S' in the keyword string, understating PRISM-RAG's factual accuracy on this instance. PRISM-RAG retrieves DC-specific statutory text via its jurisdiction-aware direct injection path (JA\,=\,1.0 for this instance), enabling it to cite the applicable DC flavored tobacco product framework verbatim. StandardRAG retrieves semantically similar \emph{federal} FDA regulatory content, which is not relevant for the question scope.
}
\label{fig:qualitative-comparison}
\end{figure*}

Figure \ref{fig:qualitative-comparison} shows one of the most dissimilar response pairs across the 1,325 policy QA instances. The question asks how Washington, D.C.\ statutes classify nicotine pouches. PRISM-RAG answers with two specific D.C.\ Code citations (\S\,7-1721.08(b) and \S\,7-1721.01(1)), found in the KG, correctly explaining that D.C.'s flavored tobacco prohibitions cover synthetic nicotine products that impart a characterizing flavor, and that a flavored nicotine pouch therefore falls under D.C.'s \textit{flavored tobacco product prohibition} rather than a separate type of nicotine product. The response is \textit{grounded} in the correct jurisdiction's statutory text from our KG, retrieved via PRISM-RAG's mandatory direct injection of D.C.\ policy document chunks, boosted because the query contains a recognized jurisdiction name.
StandardRAG, however, retrieves semantically similar but wrong-jurisdiction content. D.C.\ tobacco statutes use language nearly identical to the federal ``Family Smoking Prevention and Tobacco Control Act'', which also classifies nicotine products as tobacco products under FDA oversight. Pure semantic retrieval betrays the model response, causing it to form context irrelevant to the question. The StandardRAG response correctly identifies the federal classification but explicitly acknowledges that no D.C.-specific statutory information was retrieved, and it cannot confirm whether D.C.\ law treats nicotine pouches separately. While it is positive that retrieval failure is reported honestly, it does carry a consequence. That is, a public health researcher relying on this answer for D.C.\ compliance monitoring would receive accurate federal context and \textit{no actionable local determination}. 
This critical failure case, where chunks are semantically similar but jurisdictionally wrong in retrieval, is precisely the cross-jurisdictional ambiguity that PRISM-RAG's jurisdiction-aware $\mathcal{V}_T$ scoring boost and direct $\mathcal{V}_P$ injection are designed to resolve.

\section{Conclusions} \label{sec:conclusion}

In this work, we introduced NicoPRISM and PRISM-RAG to address inter-context conflict within the difficult problem setting of tobacco and nicotine product surveillance. 
HippoRAG2 and HyperGraphRAG, SOTA document RAG methods, achieve CA of 8.4\% and 1.3\%, respectively, compared with PRISM-RAG's 41.78\%. StandardRAG underperforms at retrieving from the correct jurisdiction in only 44.95\% of policy queries (JA), compared to PRISM-RAG's 93.87\%. This performance gap is statistically unambiguous ($p < 0.001$, 95\% CI: [46.2, 50.9]\%) and directly explains why a method that scores competitive CA can still be retrieving the wrong jurisdiction's statute in over half of all queries. HyperGraphRAG's high RG (0.885) yet near-zero CA (1.3\%) further reveals that a highly grounded response does not guarantee a proportionally high compliance determination, an instance of inter-context conflict, whereby a response is fluently grounded in retrieved evidence yet built on the wrong jurisdiction's statute. 
This discrepancy shows that JA (\ie, source determination) is a necessary condition for reliable policy compliance reasoning. PRISM-RAG addresses these reasoning limitations through image-conditioned hyperedge traversal with strong document understanding, concept hyperedges that connect marketing attribute language to statutory terminology, and jurisdiction-aware context assembly, which disambiguates semantically similar statutory language across jurisdictions.

\bibliographystyle{IEEEtran}
\bibliography{main}

@misc{noauthor_who_nodate,
	title = {{WHO} global report on trends in prevalence of tobacco use 2000–2024 and projections 2025–2030},
	url = {https://www.who.int/publications/i/item/9789240116276},
	language = {en},
	urldate = {2026-05-01},
}

@article{murthy-tobacco-dataset-2024,
  title={Using computer vision to detect e-cigarette content in TikTok videos},
  author={Murthy, Dhiraj and Ouellette, Rachel R and Anand, Tanvi and Radhakrishnan, Srijith and Mohan, Nikhil C and Lee, Juhan and Kong, Grace},
  journal={Nicotine and Tobacco Research},
  volume={26},
  number={Supplement\_1},
  pages={S36--S42},
  year={2024},
  publisher={Oxford University Press US}
}

@article{vassey2024scalable-tobacco,
  title={Scalable surveillance of E-cigarette products on Instagram and Tiktok using computer vision},
  author={Vassey, Julia and Kennedy, Chris J and Herbert Chang, Ho-Chun and Smith, Ashley S and Unger, Jennifer B},
  journal={Nicotine and Tobacco Research},
  volume={26},
  number={5},
  pages={552--560},
  year={2024},
  publisher={Oxford University Press US}
}

@article{chappa2024phad,
  title={Public health advocacy dataset: A dataset of tobacco usage videos from social media},
  author={Chappa, Naga VS and McCormick, Charlotte and Gongora, Susana Rodriguez and Dobbs, Page Daniel and Luu, Khoa},
  journal={arXiv preprint arXiv:2411.13572},
  year={2024}
}

@article{kim2021machine,
  title={Machine learning models of tobacco susceptibility and current use among adolescents from 97 countries in the Global Youth Tobacco Survey, 2013-2017},
  author={Kim, Nayoung and Loh, Wei-Yin and McCarthy, Danielle E},
  journal={PLOS Global Public Health},
  volume={1},
  number={12},
  pages={e0000060},
  year={2021},
  publisher={Public Library of Science San Francisco, CA USA}
}

@article{perski2023classification,
  title={Classification of lapses in smokers attempting to stop: A supervised machine learning approach using data from a popular smoking cessation smartphone app},
  author={Perski, Olga and Li, Kezhi and Pontikos, Nikolas and Simons, David and Goldstein, Stephanie P and Naughton, Felix and Brown, Jamie},
  journal={Nicotine and Tobacco Research},
  volume={25},
  number={7},
  pages={1330--1339},
  year={2023},
  publisher={Oxford University Press US}
}

@article{lakatos2024multimodal,
  title={A multimodal deep learning architecture for smoking detection with a small data approach},
  author={Lakatos, R{\'o}bert and Pollner, P{\'e}ter and Hajdu, Andr{\'a}s and Jo{\'o}, Tam{\'a}s},
  journal={Frontiers in Artificial Intelligence},
  volume={7},
  pages={1326050},
  year={2024},
  publisher={Frontiers Media SA}
}

@inproceedings{chappa2024advanced,
  title={Advanced deep learning techniques for tobacco usage assessment in tiktok videos},
  author={Chappa, Naga VS Raviteja and McCormick, Charlotte and Gongora, Susana Rodriguez and Dobbs, Page Daniel and Luu, Khoa},
  booktitle={2024 IEEE Green Technologies Conference (GreenTech)},
  pages={162--163},
  year={2024},
  organization={IEEE}
}

@article{kong2023understanding,
  title={Understanding e-cigarette content and promotion on YouTube through machine learning},
  author={Kong, Grace and Schott, Alex Sebastian and Lee, Juhan and Dashtian, Hassan and Murthy, Dhiraj},
  journal={Tobacco control},
  volume={32},
  number={6},
  pages={739--746},
  year={2023},
  publisher={BMJ Publishing Group Ltd}
}

@article{murthy2023influence,
  title={Influence of user profile attributes on e-cigarette--related searches on youtube: Machine learning clustering and classification},
  author={Murthy, Dhiraj and Lee, Juhan and Dashtian, Hassan and Kong, Grace and others},
  journal={JMIR infodemiology},
  volume={3},
  number={1},
  pages={e42218},
  year={2023},
  publisher={JMIR Publications Inc., Toronto, Canada}
}

@article{twitter-tobacco-llm-2025,
    author = {Küçükali, Hüseyin and Erdoğan, Mehmet Sarper},
    title = {AI for Tobacco Control: Identifying Tobacco-Promoting Social Media Content Using Large Language Models},
    journal = {Nicotine and Tobacco Research},
    volume = {27},
    number = {6},
    pages = {988-996},
    year = {2025},
    month = {06},
    issn = {1469-994X},
    doi = {10.1093/ntr/ntae276},
    url = {https://doi.org/10.1093/ntr/ntae276},
    eprint = {https://academic.oup.com/ntr/article-pdf/27/6/988/60799608/ntae276.pdf},
}

@article{vaswani2017attention,
  title={Attention is all you need},
  author={Vaswani, Ashish and Shazeer, Noam and Parmar, Niki and Uszkoreit, Jakob and Jones, Llion and Gomez, Aidan N and Kaiser, {\L}ukasz and Polosukhin, Illia},
  journal={Advances in neural information processing systems},
  volume={30},
  year={2017}
}

@article{jimenez2024hipporag, 
  title={Hipporag: Neurobiologically inspired long-term memory for large language models},
  author={Jimenez Gutierrez, Bernal and Shu, Yiheng and Gu, Yu and Yasunaga, Michihiro and Su, Yu},
  journal={Advances in Neural Information Processing Systems},
  volume={37},
  pages={59532--59569},
  year={2024}
}

@misc{gutierrez2025ragmemorynonparametriccontinual,
      title={From RAG to Memory: Non-Parametric Continual Learning for Large Language Models}, 
      author={Bernal Jiménez Gutiérrez and Yiheng Shu and Weijian Qi and Sizhe Zhou and Yu Su},
      year={2025},
      eprint={2502.14802},
      archivePrefix={arXiv},
      primaryClass={cs.CL},
      url={https://arxiv.org/abs/2502.14802}, 
}

@misc{han2025graphrag,
      title={Retrieval-Augmented Generation with Graphs (GraphRAG)}, 
      author={Haoyu Han and Yu Wang and Harry Shomer and Kai Guo and Jiayuan Ding and Yongjia Lei and Mahantesh Halappanavar and Ryan A. Rossi and Subhabrata Mukherjee and Xianfeng Tang and Qi He and Zhigang Hua and Bo Long and Tong Zhao and Neil Shah and Amin Javari and Yinglong Xia and Jiliang Tang},
      year={2025},
      eprint={2501.00309},
      archivePrefix={arXiv},
      primaryClass={cs.IR},
      url={https://arxiv.org/abs/2501.00309}, 
}

@misc{hu2025graggraphretrievalaugmentedgeneration,
      title={GRAG: Graph Retrieval-Augmented Generation}, 
      author={Yuntong Hu and Zhihan Lei and Zheng Zhang and Bo Pan and Chen Ling and Liang Zhao},
      year={2025},
      eprint={2405.16506},
      archivePrefix={arXiv},
      primaryClass={cs.LG},
      url={https://arxiv.org/abs/2405.16506}, 
}

@misc{guo2025lightrag,
      title={LightRAG: Simple and Fast Retrieval-Augmented Generation}, 
      author={Zirui Guo and Lianghao Xia and Yanhua Yu and Tu Ao and Chao Huang},
      year={2025},
      eprint={2410.05779},
      archivePrefix={arXiv},
      primaryClass={cs.IR},
      url={https://arxiv.org/abs/2410.05779}, 
}

@misc{chen2025pathrag,
      title={PathRAG: Pruning Graph-based Retrieval Augmented Generation with Relational Paths}, 
      author={Boyu Chen and Zirui Guo and Zidan Yang and Yuluo Chen and Junze Chen and Zhenghao Liu and Chuan Shi and Cheng Yang},
      year={2025},
      eprint={2502.14902},
      archivePrefix={arXiv},
      primaryClass={cs.CL},
      url={https://arxiv.org/abs/2502.14902}, 
}

@misc{luo2025hypergraphrag,
      title={HyperGraphRAG: Retrieval-Augmented Generation via Hypergraph-Structured Knowledge Representation}, 
      author={Haoran Luo and Haihong E and Guanting Chen and Yandan Zheng and Xiaobao Wu and Yikai Guo and Qika Lin and Yu Feng and Zemin Kuang and Meina Song and Yifan Zhu and Luu Anh Tuan},
      year={2025},
      eprint={2503.21322},
      archivePrefix={arXiv},
      primaryClass={cs.AI},
      url={https://arxiv.org/abs/2503.21322}, 
}

@misc{yenduri2023gpt,
      title={Generative Pre-trained Transformer: A Comprehensive Review on Enabling Technologies, Potential Applications, Emerging Challenges, and Future Directions}, 
      author={Gokul Yenduri and Ramalingam M and Chemmalar Selvi G and Supriya Y and Gautam Srivastava and Praveen Kumar Reddy Maddikunta and Deepti Raj G and Rutvij H Jhaveri and Prabadevi B and Weizheng Wang and Athanasios V. Vasilakos and Thippa Reddy Gadekallu},
      year={2023},
      eprint={2305.10435},
      archivePrefix={arXiv},
      primaryClass={cs.CL},
      url={https://arxiv.org/abs/2305.10435}, 
}

@misc{brown2020languagemodelsfewshotlearners,
      title={Language Models are Few-Shot Learners}, 
      author={Tom B. Brown and Benjamin Mann and Nick Ryder and Melanie Subbiah et al.},
      year={2020},
      eprint={2005.14165},
      archivePrefix={arXiv},
      primaryClass={cs.CL},
      url={https://arxiv.org/abs/2005.14165}, 
}

@misc{openai2024gpt4technicalreport,
      title={GPT-4 Technical Report}, 
      author={OpenAI and Josh Achiam and Steven Adler and Sandhini Agarwal et al.},
      year={2024},
      eprint={2303.08774},
      archivePrefix={arXiv},
      primaryClass={cs.CL},
      url={https://arxiv.org/abs/2303.08774}, 
}

@misc{singh2025openaigpt5card,
      title={OpenAI GPT-5 System Card}, 
      author={Aaditya Singh and Adam Fry and Adam Perelman and Adam Tart and Adi Ganesh, et al.},
      year={2025},
      eprint={2601.03267},
      archivePrefix={arXiv},
      primaryClass={cs.CL},
      url={https://arxiv.org/abs/2601.03267}, 
}

@misc{gemmateam2025gemma3technicalreport, 
      title={Gemma 3 Technical Report}, 
      author={Gemma Team and Aishwarya Kamath and Johan Ferret and Shreya Pathak and Nino Vieillard et al.},
      year={2025},
      eprint={2503.19786},
      archivePrefix={arXiv},
      primaryClass={cs.CL},
      url={https://arxiv.org/abs/2503.19786}, 
}

@misc{anthropic2024claude3,
  title={The Claude 3 Model Family: Opus, Sonnet, Haiku},
  author={Anthropic},
  year={2024},
  eprint={2404.XXXXX},
  archivePrefix={arXiv},
  primaryClass={cs.CL}
}

@misc{ouyang2022traininglanguagemodelsfollow,
      title={Training language models to follow instructions with human feedback}, 
      author={Long Ouyang and Jeff Wu and Xu Jiang and Diogo Almeida and Carroll L. Wainwright and Pamela Mishkin and Chong Zhang and Sandhini Agarwal and Katarina Slama and Alex Ray and John Schulman and Jacob Hilton and Fraser Kelton and Luke Miller and Maddie Simens and Amanda Askell and Peter Welinder and Paul Christiano and Jan Leike and Ryan Lowe},
      year={2022},
      eprint={2203.02155},
      archivePrefix={arXiv},
      primaryClass={cs.CL},
      url={https://arxiv.org/abs/2203.02155}, 
}

@misc{stiennon2022learningsummarizehumanfeedback,
      title={Learning to summarize from human feedback}, 
      author={Nisan Stiennon and Long Ouyang and Jeff Wu and Daniel M. Ziegler and Ryan Lowe and Chelsea Voss and Alec Radford and Dario Amodei and Paul Christiano},
      year={2022},
      eprint={2009.01325},
      archivePrefix={arXiv},
      primaryClass={cs.CL},
      url={https://arxiv.org/abs/2009.01325}, 
}

@article{peng2023instruction,
  title={Instruction Tuning with GPT-4},
  author={Peng, Baolin and Li, Chunyuan and He, Pengcheng and Galley, Michel and Gao, Jianfeng},
  journal={arXiv preprint arXiv:2304.03277},
  year={2023}
}

@misc{touvron2023llama2openfoundation,
      title={Llama 2: Open Foundation and Fine-Tuned Chat Models}, 
      author={Hugo Touvron and Louis Martin and Kevin Stone and Peter Albert et al.},
      year={2023},
      eprint={2307.09288},
      archivePrefix={arXiv},
      primaryClass={cs.CL},
      url={https://arxiv.org/abs/2307.09288}, 
}

@misc{touvron2023llamaopenefficientfoundation,
      title={LLaMA: Open and Efficient Foundation Language Models}, 
      author={Hugo Touvron and Thibaut Lavril and Gautier Izacard and Xavier Martinet and Marie-Anne Lachaux and Timothée Lacroix and Baptiste Rozière and Naman Goyal and Eric Hambro and Faisal Azhar and Aurelien Rodriguez and Armand Joulin and Edouard Grave and Guillaume Lample},
      year={2023},
      eprint={2302.13971},
      archivePrefix={arXiv},
      primaryClass={cs.CL},
      url={https://arxiv.org/abs/2302.13971}, 
}

@misc{liu2023visualinstructiontuning,
      title={Visual Instruction Tuning}, 
      author={Haotian Liu and Chunyuan Li and Qingyang Wu and Yong Jae Lee},
      year={2023},
      eprint={2304.08485},
      archivePrefix={arXiv},
      primaryClass={cs.CV},
      url={https://arxiv.org/abs/2304.08485}, 
}

@misc{liu2024improvedbaselinesvisualinstruction,
      title={Improved Baselines with Visual Instruction Tuning}, 
      author={Haotian Liu and Chunyuan Li and Yuheng Li and Yong Jae Lee},
      year={2024},
      eprint={2310.03744},
      archivePrefix={arXiv},
      primaryClass={cs.CV},
      url={https://arxiv.org/abs/2310.03744}, 
}

@misc{zhang2025llavaminiefficientimagevideo,
      title={LLaVA-Mini: Efficient Image and Video Large Multimodal Models with One Vision Token}, 
      author={Shaolei Zhang and Qingkai Fang and Zhe Yang and Yang Feng},
      year={2025},
      eprint={2501.03895},
      archivePrefix={arXiv},
      primaryClass={cs.CV},
      url={https://arxiv.org/abs/2501.03895}, 
}

@misc{wu2025qwenimagetechnicalreport,
      title={Qwen-Image Technical Report}, 
      author={Chenfei Wu and Jiahao Li and Jingren Zhou and Junyang Lin and Kaiyuan Gao and Kun Yan and Sheng-ming Yin and Shuai Bai and Xiao Xu and Yilei Chen and Yuxiang Chen and Zecheng Tang and Zekai Zhang and Zhengyi Wang and An Yang and Bowen Yu and Chen Cheng and Dayiheng Liu and Deqing Li and Hang Zhang and Hao Meng and Hu Wei and Jingyuan Ni and Kai Chen and Kuan Cao and Liang Peng and Lin Qu and Minggang Wu and Peng Wang and Shuting Yu and Tingkun Wen and Wensen Feng and Xiaoxiao Xu and Yi Wang and Yichang Zhang and Yongqiang Zhu and Yujia Wu and Yuxuan Cai and Zenan Liu},
      year={2025},
      eprint={2508.02324},
      archivePrefix={arXiv},
      primaryClass={cs.CV},
      url={https://arxiv.org/abs/2508.02324}, 
}

@misc{bge-m3_2024,
      title={BGE M3-Embedding: Multi-Lingual, Multi-Functionality, Multi-Granularity Text Embeddings Through Self-Knowledge Distillation}, 
      author={Jianlv Chen and Shitao Xiao and Peitian Zhang and Kun Luo and Defu Lian and Zheng Liu},
      year={2024},
      eprint={2402.03216},
      archivePrefix={arXiv},
      primaryClass={cs.CL}
}

@misc{zhai2023sigmoidlosslanguageimage,
      title={Sigmoid Loss for Language Image Pre-Training}, 
      author={Xiaohua Zhai and Basil Mustafa and Alexander Kolesnikov and Lucas Beyer},
      year={2023},
      eprint={2303.15343},
      archivePrefix={arXiv},
      primaryClass={cs.CV},
      url={https://arxiv.org/abs/2303.15343}, 
}

@misc{apify-citation,
  author = {Apify},
  title = {Apify Technologies, Copyright @2024},
  year = 2024,
  url = {http://apify.com},
  urldate = {2024-11-05}
}

@article{huang2025survey-hallu,
  title={A survey on hallucination in large language models: Principles, taxonomy, challenges, and open questions},
  author={Huang, Lei and Yu, Weijiang and Ma, Weitao and Zhong, Weihong and Feng, Zhangyin and Wang, Haotian and Chen, Qianglong and Peng, Weihua and Feng, Xiaocheng and Qin, Bing and others},
  journal={ACM Transactions on Information Systems},
  volume={43},
  number={2},
  pages={1--55},
  year={2025},
  publisher={ACM New York, NY}
}

@article{liu2024survey-hallu,
  title={A survey on hallucination in large vision-language models},
  author={Liu, Hanchao and Xue, Wenyuan and Chen, Yifei and Chen, Dapeng and Zhao, Xiutian and Wang, Ke and Hou, Liping and Li, Rongjun and Peng, Wei},
  journal={arXiv preprint arXiv:2402.00253},
  year={2024}
}

@article{que2024hellobench,
  title={Hellobench: Evaluating long text generation capabilities of large language models},
  author={Que, Haoran and Duan, Feiyu and He, Liqun and Mou, Yutao and Zhou, Wangchunshu and Liu, Jiaheng and Rong, Wenge and Wang, Zekun Moore and Yang, Jian and Zhang, Ge and others},
  journal={arXiv preprint arXiv:2409.16191},
  year={2024}
}

\end{document}